\documentclass[lettersize,journal]{IEEEtran}
\usepackage{enumerate}
\usepackage[nottoc]{tocbibind}
\usepackage{enumitem,kantlipsum}
\usepackage{indentfirst}
\usepackage{multirow}
\usepackage{pifont}
\usepackage[table,xcdraw,dvipsnames,usenames]{xcolor}
\usepackage{subfloat}
\usepackage[noadjust]{cite} 
\usepackage{amsmath,amssymb,amsfonts}
\usepackage{algorithm}
\usepackage{algpseudocode}
\usepackage{caption}
\usepackage{subcaption}
\DeclareCaptionLabelFormat{ieeetable}{\MakeUppercase{#1~#2}\\}
\usepackage{textcomp}
\usepackage{stfloats}
\usepackage{url}
\usepackage{graphicx}
\usepackage{bbding}
\usepackage{rotating}
\usepackage{gensymb}
\usepackage[bottom,flushmargin]{footmisc}
\usepackage{adjustbox}
\usepackage{xcolor}
\usepackage{siunitx}
\usepackage{lineno}
\usepackage{multicol,siunitx,amssymb,enumitem,ragged2e, pifont}
\usepackage{hyperref}
\usepackage{stmaryrd}
\usepackage{trimclip}
\newcolumntype{K}[1]{>{\centering\arraybackslash\scriptsize}p{#1}}
\newlength{\kw}
\def\BibTeX{{\rm B\kern-.05em{\sc i\kern-.025em b}\kern-.08em
    T\kern-.1667em\lower.7ex\hbox{E}\kern-.125emX}}
\usepackage{balance}
\usepackage{xr}
\begin{document}
\title{
Low-Latency Video Anonymization for Crowd Anomaly Detection: Privacy Versus Performance
}

\author{Mulugeta~Weldezgina~Asres,
        Lei~Jiao, \textit{Senior Member, IEEE},
        Christian~Walter~Omlin
\thanks{This work was funded by the AI4Citizens project (ID: 320783, and project correspondence: L.~Jiao), under the Research Council of Norway.}
\thanks{M.~W.~Asres is with University~of~Agder, Norway, mulugetawa@uia.no.}   
\thanks{L.~Jiao is with University~of~Agder, Norway, lei.jiao@uia.no 
}   
\thanks{C.~W.~Omlin is with University~of~Agder, Norway, christian.omlin@uia.no.}   
 }

\markboth{IEEE Transactions on Information Forensics and Security}%
{Author \MakeLowercase{\textit{et al.}}: Bare Demo of IEEEtran.cls for IEEE Journals}

\maketitle

\begin{abstract}
Recent advancements in artificial intelligence hold ample potential for monitoring applications using surveillance cameras. However, concerns about privacy and model bias have made it challenging to utilize them in public. Although de-identification approaches have been proposed in the literature, aiming to achieve a certain level of anonymization (AN), most of them employ deep learning models that are computationally demanding for real-time edge deployment. This study revisits conventional AN solutions for privacy protection and real-time video anomaly detection (VAD) applications. We propose a lightweight adaptive AN for VAD (LA3D) that employs dynamic adjustment to enhance full-body privacy protection. We have evaluated privacy protection and VAD utility retention efficacy using several publicly available datasets to examine the strengths and weaknesses of different AN methods and highlight the promising leverage of our approach. Our experiment demonstrates that the LA3D enables substantial improvement in privacy AN without severely degrading VAD efficacy, outperforming conventional and deep learning approaches.
Code: \href{https://github.com/muleina/LA3D}{https://github.com/muleina/LA3D}

\end{abstract}


\section*{{Acronyms}} 

\newlength\mylen
\settowidth\mylen{\textsc{Market15}} 
\setlist[description]{style=nextline,      
                      font==\tiny,          
                      leftmargin=5.5em,       
                      labelsep=0.5em          
                    }
\begin{description}[font=\normalfont\space\small, labelwidth=\mylen]
    \item[AN]              \small Anonymization  
    \item[AUC]             Area under the Curve    
    \item[CMC]             Cumulative Matching Characteristics Curve 
    \item[CMC-Rk]          CMC of the Top Rank k\% 
    \item[CV]              Computer Vision 
    \item[DL]              Deep Learning 
    \item[LA3D]            Our Lightweight Adaptive VA for VAD 
    \item[mAP]             Mean Average Precision 
    \item[Market1501]     Cross-Camera ReID Benchmark Dataset 
    \item[MGFN]            Magnitude-Contrastive Glance-and-Focus Network 
    \item[mR]            Mean Recall 
    \item[PD]              Privacy Attribute Detection  
    \item[PEL4VAD]         Prompt-Enhanced Learning for VAD   
    \item[ReID]            Person Re-Identification 
    \item[UCF-Crime]              VAD Benchmark Crime Dataset  
    \item[VISPR]            PD Benchmark Dataset 
    \item[VAD]               Video Anomaly Detection  
    \item[WSAD]             Weakly Supervised VAD 
    \item[XD]              VAD Benchmark Violence Dataset
    \item[$\mathbf{c}$]       Set of Privacy Attribute Classes 
    \item[$H$, $W$]            Image Height, Image Width 
    \item[$h$, $w$]             Subscript for $H$, $W$  
     \item[$\text{0}$, $\text{a}$]             Superscript for Non-Adaptive AN, Adaptive AN  
      \item[$\mathbf{z}$, $\mathbf{z_\text{ref}}$]            Image Size $\mathbf{z}=[H, W]$, Reference Image Size
    \item[$\mathbf{d}$]     Downsizing Factor of $\mathcal{P}$,  $\mathbf{d}=[d_h, d_w]$           
     \item[$\mathbf{k}$]     Kernel Window Size of $\mathcal{G}$,  $\mathbf{k}=[k_h, k_w]$               
     \item[$\boldsymbol{\sigma}$]     Kernel Standard Deviation of $\mathcal{G}$,  $\boldsymbol{\sigma}=[\sigma_h, \sigma_w]$               
    \item[$\mathbf{X}$, $\hat{\mathbf{X}}$]               Video Data, Anonymized Video Data
    \item[$\mathbf{I}$, $\hat{\mathbf{I}}$]              Frame Image, Anonymized Frame 
     \item[$\mathbf{M}$]  Binary Segmentation Mask/s
      \item[$\Psi$]      Object Detection and Segmentation Model 
    \item[$\Theta$]                Image AN Function / Algorithm
    \item[$\Theta^\text{0}$,    $\Theta^\text{a}$ ]                Non-Adaptive $\Theta$, Adaptive $\Theta$ 
    \item[$\mathcal{F}_\Theta$]             Video AN System that Applies $\Theta$ 
    \item[$U$, $P$]             VAD Utility Task, Privacy Detection Task
    \item[$\mathcal{F}_U$, $\mathcal{F}_P$]             VAD Model, PD Model
    \item[$\Gamma$]       Anonymized VAD system 
     \item[$\mathcal{G}_{\mathbf{k},\boldsymbol{\sigma}}$, $\mathcal{G}$]     Gaussian Blurring $\Theta$ with Parameter $\mathbf{k}$ and $\boldsymbol{\sigma}$    
      \item[$\mathcal{P}_\mathbf{d}$, $\mathcal{P}$]     Pixelization $\Theta$ with Parameter $\mathbf{d}$      
\end{description}

 \section{Introduction}
\label{sec:introduction}

\IEEEPARstart{P}{rivacy} and safety security are the cornerstones of a thriving society and are often intricately linked to one another~\cite{orekondy2017towards, liu2025privacy}. 
Robust security measures can safeguard privacy, and respect for privacy can foster responsible security practices and accountability.  
Striking the right balance between privacy and security is essential but an ongoing challenge~\cite{lee2024balancing}. 
With over a billion surveillance cameras worldwide, there has been a growing interest in privacy-preserving computer vision (CV) systems from industry, academia, and regulatory bodies~\cite{wu2018towards, meden2021privacy}. 
Artificial intelligence (AI) in crowd monitoring for safety presents both advantages and multifaceted challenges in the interplay between privacy and security~\cite{fioresi2023ted}. 
Several studies have demonstrated the existing concerns of deep learning (DL) models, including but not limited to a chilling effect on free movement and expression, bias, risk of profiling, and discrimination~\cite{bender2021dangers, fioresi2023ted}.
Visual anonymization (AN) obscures people's biometric identities to mitigate unintentional and malicious risks associated with AI tools, which enforces the AI monitoring to focus on behavior, not identity~\cite{meden2021privacy, dave2022spact,fioresi2023ted}. 

Biometric privacy-protecting monitoring systems aim to anonymize data while maintaining utilities relevant to the downstream applications when the privacy and utility are not necessarily trade-offs~\cite{meden2021privacy, birnstill2015user, ren2018learning, lee2024balancing}. 
Several techniques have been investigated for AN, such as conventional algorithms: masking, pixelization, and blurring~\cite{angus2022real, raina2023egoblur}, and DL-based image synthesis: face and body inpainting~\cite{brkic2017know, ren2018learning, hukkelaas2019deepprivacy, nousi2020deep, maximov2020ciagan, li2021deepblur, zhai2022a3gan, chu2023medm, hukkelaas2023deepprivacy2} or skeleton extraction~\cite{su2023prime}, and utility aware image-level obfuscation~\cite{wu2018towards, wu2020privacy, dave2022spact, fioresi2023ted}. 
Deep learning approaches enable end-to-end automation of removing privacy-sensitive attributes while preserving task utility~\cite{wu2018towards, wu2020privacy, nousi2020deep, dave2022spact, fioresi2023ted, cai2023disguise}. However, the methods require either retraining the downstream task model on the new anonymized footage or the AN becomes computationally expensive.

Visual AN tools can be deployed on an edge (camera) or cloud systems. Cloud AN systems offer broader access to resources but require substantial infrastructure investments. 
In contrast, besides addressing the increasing concern of sharing private information with the cloud, edge AN can provide further protection against privacy troubles that arise from unsecured channels between the cloud and the camera~\cite{wu2018towards}.
Although generative DL models are promising in preserving realistic image quality, they are computationally demanding, which often challenges resource-limited edge devices~\cite{maximov2020ciagan, chu2023medm, hukkelaas2023deepprivacy2}.  
Conventional AN approaches are computationally preferable but struggle to balance AN and image quality for downstream tasks. Heavy AN leads to elevated privacy protection but can also severely compromise image utility and vice versa~\cite{hukkelaas2019deepprivacy, wu2020privacy, dave2022spact, fioresi2023ted, lee2024balancing}. 
In addition, the efficacy of the AN heavily depends on the hyperparameters of the AN method. 
Despite the broad deployment of such techniques in the real world, the literature barely explores optimizing their hyperparameters besides being utilized as benchmarking baselines~\cite{ren2018learning, hukkelaas2019deepprivacy, wu2020privacy, dave2022spact, fioresi2023ted, hukkelaas2023deepprivacy2, lee2024balancing}. 
Manually adjusting the hyperparameters to anonymize a few images, such as those in news reports, may not pose a significant challenge. However, automated parameter adaptation is necessary to apply these techniques to streaming videos, such as CCTV. 

In this paper, we explore and enhance lightweight AN methods for real-time CCTV crime monitoring through video anomaly detection (VAD) applications. 
We attempt to address the potential issue of the conventional AN algorithms that arise from utilizing fixed parameters to process images, i.e., inadequate AN and severe image quality degradation. 
Fixed AN parameters are not optimized to handle the target subjects in images that differ in size or are positioned at varying depths. To address the above issue, we introduce a {lightweight adaptive AN for VAD (\textsc{LA3D})} that adjusts its hyperparameters dynamically based on the size and depth variation of the subjects in a video frame to enhance AN. 
Our approach emphasizes safeguarding a person's entire body. Thus, it detects human subjects using a fast segmentation model and applies dynamic obfuscation based on the surface area attributes of each subject.

We have conducted experiments on different publicly available datasets to evaluate the competency of privacy protection using recognition attack DL models for privacy attribute detection and person re-identification. We have also assessed the utility quality after AN using weakly supervised VAD models on large-scale video datasets. Moreover, we present a computational cost comparison and visual quality assessment. We provide detailed result discussions and analyses using both qualitative and quantitative presentations.
Our experiments demonstrate that \textsc{LA3D} significantly improves the privacy AN without apparent degradation in VAD, and outperforms DL-based approaches in accuracy and computational efficiency.
The key contributions of this study are summarized as follows:
\begin{itemize}
    \item We revisit lightweight AN methods and elaborate on their effectiveness in protecting privacy and enabling VAD. 
    \item We present a baseline performance report in privacy attribute leakage detection, person re-identification, VAD, cost analysis, and video embedding quality retention.
    \item We propose an adaptive and computationally efficient AN approach that enhances privacy protection through dynamic AN without sacrificing the VAD quality. The approach outperforms state-of-the-art DL approaches on AN, computation efficiency, and VAD accuracy.
    \item We improve a privacy attribute detector model augmented by a pre-trained encoder and weighted cost function.
\end{itemize}

The remainder of the paper is organized as follows. We briefly review related AN and VAD studies in Section~\ref{sec:background}. Thereafter, we present our methods and evaluate their performance in Section~\ref{sec:methodology} and Section~\ref{sec:results_and_discussion}, respectively, before concluding the study in Section~\ref{sec:conclusion}. We present supplementary results and discussion in the Appendix.

\section{Literature Review}
\label{sec:background}

\begin{figure}[htbp]
\centering
\setlength{\kw}{1.8cm}
\begin{subfigure}[]{1\columnwidth}
\centering
\begin{tabular}{K{\kw}K{\kw}K{\kw}K{\kw}}
No-AN & MASKED & PIXELIZED & BLURRED
\end{tabular}
\includegraphics[width=1\linewidth]{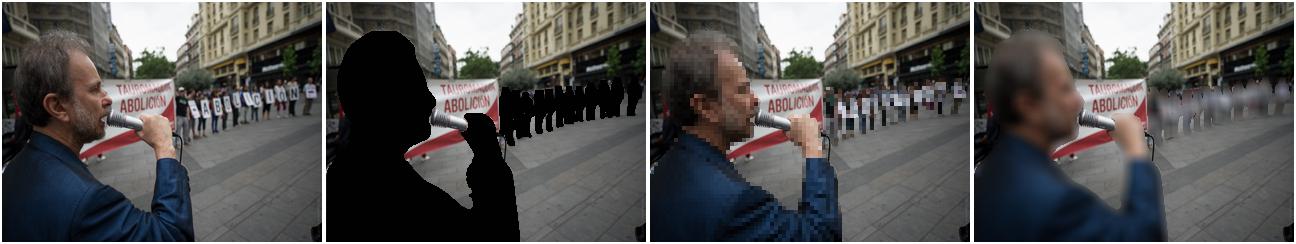} \\
\end{subfigure}
\caption{Popular conventional AN techniques on sample images from the VISPR dataset~\cite{orekondy2017towards}.}
\label{fig:vispr_anony_compare_non_adpative}
\end{figure}

Visual AN approaches strive to anonymize image or video data while maintaining relevant utilities for CV applications. 
The approaches can be categorized into different groups based on the characteristics of the target privacy attribute and the AN technique~\cite{liu2025privacy}. 
The target privacy attributes are associated with face~\cite{ren2018learning, hukkelaas2019deepprivacy, nousi2020deep, maximov2020ciagan, li2021deepblur, zhai2022a3gan, zhou2020personal}, full-body~\cite{brkic2017know, chu2023medm, hukkelaas2023deepprivacy2, su2023prime}, or entire image~\cite{wu2018towards, wu2020privacy, dave2022spact, fioresi2023ted}. The approaches may employ conventional algorithms, such as masking, pixelization, and blurring~\cite{angus2022real, raina2023egoblur, zhou2020personal}, or DL for inpainting~\cite{brkic2017know, ren2018learning, hukkelaas2019deepprivacy, nousi2020deep, maximov2020ciagan, li2021deepblur, zhai2022a3gan, chu2023medm, hukkelaas2023deepprivacy2}, skeleton extraction~\cite{su2023prime}, and utility-aware image-level obfuscation~\cite{wu2018towards, wu2020privacy, dave2022spact, fioresi2023ted, aslam2025balancing}.  
The conventional AN techniques are lightweight and promising for real-time processing and are claimed to be effective in removing privacy-sensitive information. However, they may severely alter the original data, resulting in a considerable loss in quality and rendering the generated anonymized suitability for downstream CV tasks~\cite{ren2018learning, wu2018towards, wu2020privacy, su2023prime}. They can also suffer from considerable privacy leakage due to non-optimized hyperparameters (see Fig.~\ref{fig:vispr_anony_compare_non_adpative}). 
Machine learning approaches have been proposed to remove privacy-sensitive features while preserving task utility for various CV tasks, such as action recognition~\cite{wu2018towards, ren2018learning, wu2020privacy, dave2022spact, aslam2025balancing}, VAD~\cite{fioresi2023ted, su2023prime}, non-private facial characteristics recognition~\cite{cai2023disguise}, and object detection~\cite{lee2024balancing}. 
However, these approaches require fine-tuning of the downstream task model during or after the AN model training on the new anonymized data.
The face or body inpainting generative models aim to reduce perturbation and preserve realistic image quality so that downstream tasks are less impacted, alleviating the need to retrain the task models~\cite{ren2018learning, hukkelaas2019deepprivacy, maximov2020ciagan, li2021deepblur, zhai2022a3gan, chu2023medm, hukkelaas2023deepprivacy2}.
Nevertheless, the generative approaches require intensive computation and are often much slower for real-time deployment~\cite{hukkelaas2023deepprivacy2}.
Zhao et al.~\cite{zhao2025visual} explore various visual AN methods in-depth, discussing recent studies in the field.

Crowd VAD has garnered significant attention as an essential aspect of CV in the realm of smart cities~\cite{liu2025privacy, abdalla2025video}. Many innovative DL techniques have been introduced, consistently demonstrating superior VAD~\cite{sharif2022deep}.
Unsupervised and semi-supervised VAD models dominate the arena. These models, such as autoencoders, are trained on a sequence of video frames to learn what constitutes normal activity~\cite{xie2025distributed}. Any deviation from these patterns, detected using reconstruction and prediction errors, is considered an anomaly~\cite{abdalla2025video}.
Semi-supervised VAD models often encounter overgeneralization challenges and struggle to distinguish normal characteristics from actual anomalies. As a result, recent research focuses on weakly supervised VAD (WSAD) that involves guiding the model training process using a limited set of labeled anomaly data to address overgeneralization~\cite{tian2021weakly, pu2024learning, chen2023mgfn, sharif2023cnn, sharif2023deep, liu2022decouple}. 
Moreover, VAD is bolstered by leveraging features extracted from encoders trained under natural language supervision~\cite{joo2023clip, pu2024learning, sharif2023cnn}. 
Refs.~\cite{pu2024learning, sharif2023cnn} integrate CLIP~\cite{radford2021learning} to extract discerning representations during model training and improve performance effectively. Liu et al.~\cite{liu2025privacy} present an elaborated discussion on recent privacy-preserving VAD approaches and their taxonomies.

In our experiment, we employ the state-of-the-art WSAD models, such as the prompt-enhanced learning for VAD (\textsc{PEL4VAD})~\cite{pu2024learning}, to measure the VAD utility and detect anomalies in video datasets after AN.
The \textsc{PEL4VAD} employs temporal context aggregation and augmented semantic discriminability through prompt-enhanced learning to significantly improve the computational efficiency and accuracy of WSAD, respectively~\cite{pu2024learning}. 
The temporal context aggregation efficiently captures temporal relations across video snippets by reusing a similarity matrix, reducing computational load and parameter count. The prompt-enhanced learning is achieved through knowledge-based prompts (using \textsc{ConceptNet}~\cite{speer2017conceptnet}), context separation (embedding using \textsc{CLIP}~\cite{radford2021learning}), and cross-modal alignment, enriching the model's semantic discrimination.
We also utilize other top-performing WSAD models, the magnitude-contrastive glance-and-focus network (\textsc{MGFN})~\cite{chen2023mgfn}. 
The \textsc{MGFN} utilizes transformers to glance at the whole video globally and then steer attention to each video portion, imitating the global-to-local vision system of human beings for VAD~\cite{chen2023mgfn}.
\section{Methodology}
\label{sec:methodology}

This section presents the problem formulation, the proposed approaches, and the tools employed in our study.

\subsection{Problem Formulation}

\begin{figure}[!t]
\centering
\begin{subfigure}[]{1\columnwidth}
\centering
\includegraphics[width=0.75\columnwidth]{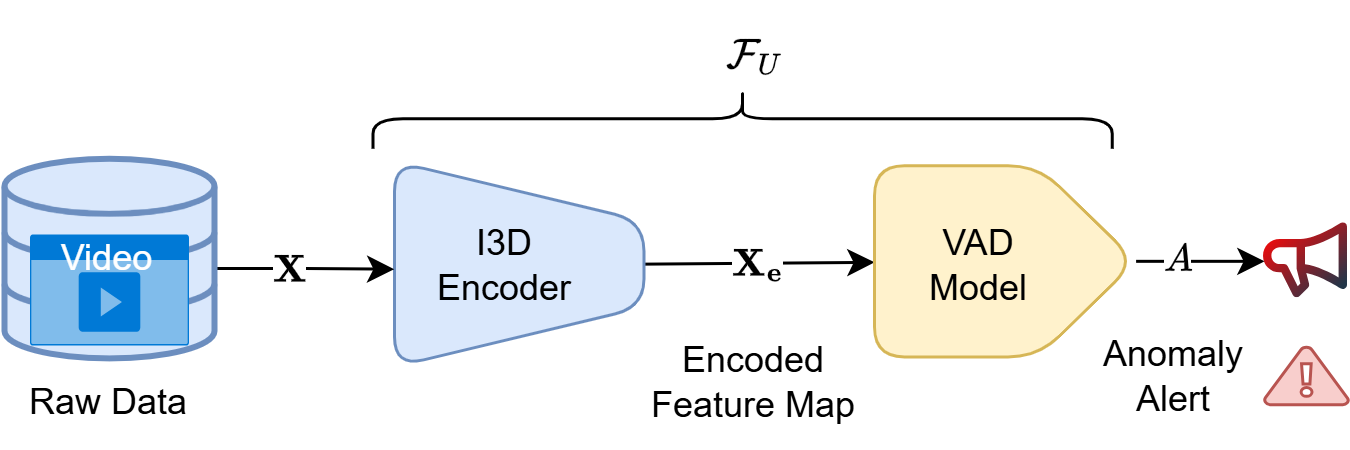} 
\caption{}
\label{fig:vad_pipe_diags_nonanonymized}
\end{subfigure}
\begin{subfigure}[]{1\columnwidth}
\centering
\includegraphics[width=1\columnwidth]{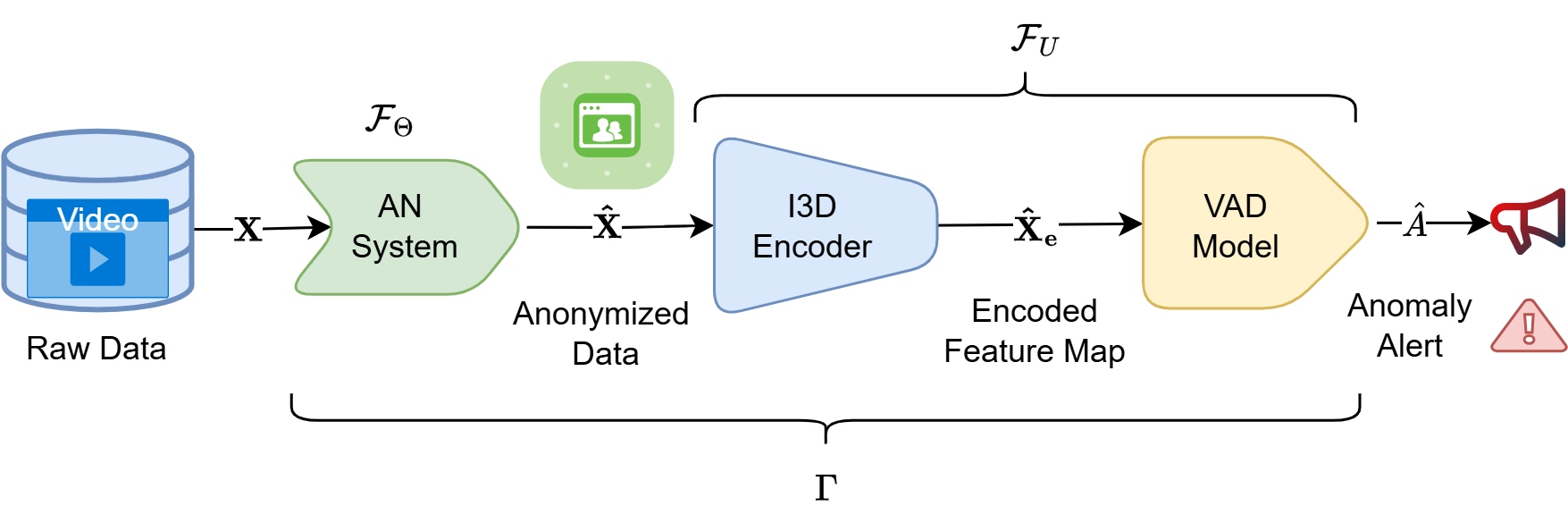}
\caption{}
\label{fig:vad_pipe_diags_anonymized}
\end{subfigure}
\caption{System pipeline diagram of the VAD systems: a) non-anonymized VAD ($\mathcal{F}_U$), and b) anonymized (privacy-preserving) VAD ($\Gamma$).}
\label{fig:vad_pipe_diags}
\vspace*{-0.5\baselineskip}
\end{figure}

This study aims to improve the privacy obfuscation and VAD utility preservation capabilities of conventional image AN techniques (see Fig.~\ref{fig:vad_pipe_diags}). 
Let’s consider a raw video data $\mathbf{X} \in \mathbb{Z^+}$, a utility VAD task $U$, and a privacy leakage task $Q$. 
The $U$ is maximized, without regard to the $Q$, for a non-anonymized VAD system $\mathcal{F}_U$ as (see in Fig.~\ref{fig:vad_pipe_diags_nonanonymized}): 
\begin{equation}
    A = \mathcal{F}_U (\mathbf{X}) : \arg \underset{{U}}{\max} ~\mathcal{F}_U (\mathbf{X}).
\end{equation}
The $\mathcal{F}_U$ process $\mathbf{X}$ to trigger a VAD alert score $A$ without protection of sensitive private information in the video, which violates the global regulations such as the GDPR and the EU AI Act~\cite{wu2018towards, meden2021privacy, fioresi2023ted, wu2020privacy}. The VAD system can exhibit bias towards specific racial, gender, age, religion, and other privacy attributes due to a lack of balanced and curated training datasets~\cite{bender2021dangers, fioresi2023ted}. 
In contrast, the goal of a privacy-preserving VAD system $\Gamma$, shown in Fig.~\ref{fig:vad_pipe_diags_anonymized}, is to maintain the efficacy of $\mathcal{F}_U$ while reducing $Q$ by integrating an AN system $\mathcal{F}_{\Theta}$ as: 
\begin{equation}
    \hat{A} = \Gamma(\mathbf{X}) : \arg \underset{{\phi}}{\max} ~\mathcal{F}_U \left(\mathcal{F}_{\Theta}(\mathbf{X})\right),
\end{equation}
where the $\hat{A}$ is the privacy-preserving VAD alert score and $\phi = \arg \underset{Q\downarrow}{\min} \arg \underset{U\uparrow}{\max}$ is the aggregate task to maximize. 
The AN system $\mathcal{F}_{\Theta}$ can be formulated as:
\begin{equation}
\label{eq:generic_an}
     \mathbf{\hat{X}} = \mathcal{F}_{\Theta} (\mathbf{X}) : \Theta (\mathbf{I}), \text{~for}~\forall \mathbf{I} \in \mathbf{X},
\end{equation}
{where $\mathbf{I}$ is a frame in the video data $\mathbf{X}$, $\mathbf{\hat{X}}$ is the anonymized version of $\mathbf{X}$, and $\Theta$ is the AN algorithm.}

Our study focuses on full-body AN that applies $\Theta$ only on the most privacy-sensitive content---the human subjects as shown in Fig.~\ref{fig:an_pipe_diags}. Thus, the above equation can be reformulated using a masking mechanism as: 
\begin{equation}
\label{eq:an_non_adaptive}
    \mathbf{\hat{X}} = \mathcal{F}_{\Theta} (\mathbf{X}) : \mathbf{I} \odot  (1 - \mathbf{M}) + \mathbf{M} \odot  \Theta (\mathbf{I}), \text{~for}~\forall \mathbf{I} \in \mathbf{X},
\end{equation}
{where $\mathbf{M}$ is the binary segmentation mask of the target objects, and $\odot$ represents element-wise multiplication. The $\mathbf{M}$ is generated using an object segmentation model $\Psi$ as:}
\begin{equation}
    \mathbf{M} = \Psi(\mathbf{I}).
\end{equation}

Although conventional AN algorithms are lightweight and promising for real-time processing, they often struggle to achieve $\phi$ (see Fig.~\ref{fig:vispr_anony_compare_non_adpative}). Heavier ANs achieve a lower $Q$ but can also deteriorate the quality of $U$ and vice versa~\cite{hukkelaas2019deepprivacy, wu2020privacy, dave2022spact, fioresi2023ted}. The AN highly depends on $\Theta$'s hyperparameters.
\begin{itemize}
\item {The Gaussian blurring $\Theta (\mathbf{I})$ can be formulated as:}
\begin{equation}
\label{eq:gaussian_kernel}
 \begin{aligned}
    &\mathbf{\hat{I}} = \mathcal{G}_{\mathbf{k},\boldsymbol{\sigma}}(\mathbf{I}):\mathbf{I} \otimes \mathbf{G},~\text{for}~\mathbf{k},\boldsymbol{\sigma} \in \mathbb{Z^+},
\end{aligned}
\end{equation}
{Here $\mathbf{\hat{I}}$ is the anonymized version of $\mathbf{I}$. The $\mathcal{G}_{\mathbf{k},\boldsymbol{\sigma}}$ is the Gaussian blurring function that smooths the image using a 2D kernel filter $\mathbf{G}$, where the $\mathbf{k}=[k_h, k_w]$ and $\boldsymbol{\sigma}=[\sigma_h, \sigma_w]$ denote the kernel's window size and standard deviation, respectively. The $\otimes$ denotes a sliding window filtering operation. The filter coefficients of $\mathbf{G}$, for $i = [0, \dots, k_h-1]$ and $j = [0, \dots, k_w-1]$, are estimated as: }
\begin{equation}
\label{eq:gaussian_kernel_coeff}
    g_{i,j} = \beta \Phi(i,h) \Phi(j,w),~\Phi(b,z) = e^{\frac{-(b-(k_z-1)/2)^2}{2\sigma_z^2}},
\end{equation}
{where $g_{i,j} \in \mathbf{G}$ holds the $(i,j)^{\text{th}}$ coefficient of the $\mathbf{G}$.}
{The $\beta$ is a factor to enable a symmetric $g$, i.e., $\sum_{i,j}^{}g_{i,j} = 1$. }

\item {The pixelization $\Theta (\mathbf{I})$ can be formulated as:}
\begin{equation}
\label{eq:pixelize_fn}
    \mathbf{\hat{I}} = \mathcal{P}_{\mathbf{d}} (\mathbf{I}): \mathcal{S_{\text{up}}} (\mathcal{S_{\text{down}}}(\mathbf{I}, \mathbf{d}), \mathbf{z}),~\text{for} ~\mathbf{d} \in \mathbb{Z^+},
\end{equation}
{where $\mathcal{P}_{\mathbf{d}}$ is a pixelization function that linearly downsamples ($\mathcal{S_{\text{down}}}$) the image $\mathbf{I}$, with the original size of $\mathbf{z}=| \mathbf{I}| =[H, W]$, by a factor of $\mathbf{d}=[d_h, d_w]$ and then upsamples ($\mathcal{S_{\text{up}}}$) it back to the original size using interpolation to nearest pixel algorithm.}
\end{itemize}

Despite the popularity, the literature has yet to explore optimizing the hyperparameters of conventional AN algorithms~\cite{hukkelaas2019deepprivacy, wu2020privacy, dave2022spact, fioresi2023ted}. 
The fixed hyperparameter settings of $\mathcal{G}_{\mathbf{k},\boldsymbol{\sigma}}$ and $\mathcal{P}_{\mathbf{d}}$ lead to varying transformation effects when the target objects in an image exhibit different sizes or depths.
Thus, we present in this study an adaptive AN that considers the potential size variation of the target objects to dynamically adjust the hyperparameters and improve the $\phi$ by minimizing $Q$ while simultaneously maintaining $U$.

\subsection{Lightweight Adaptive Full-Body Anonymization}

\begin{figure*}[!t]
\centering
\begin{subfigure}[]{0.7\textwidth}
\centering
\includegraphics[width=1\textwidth]{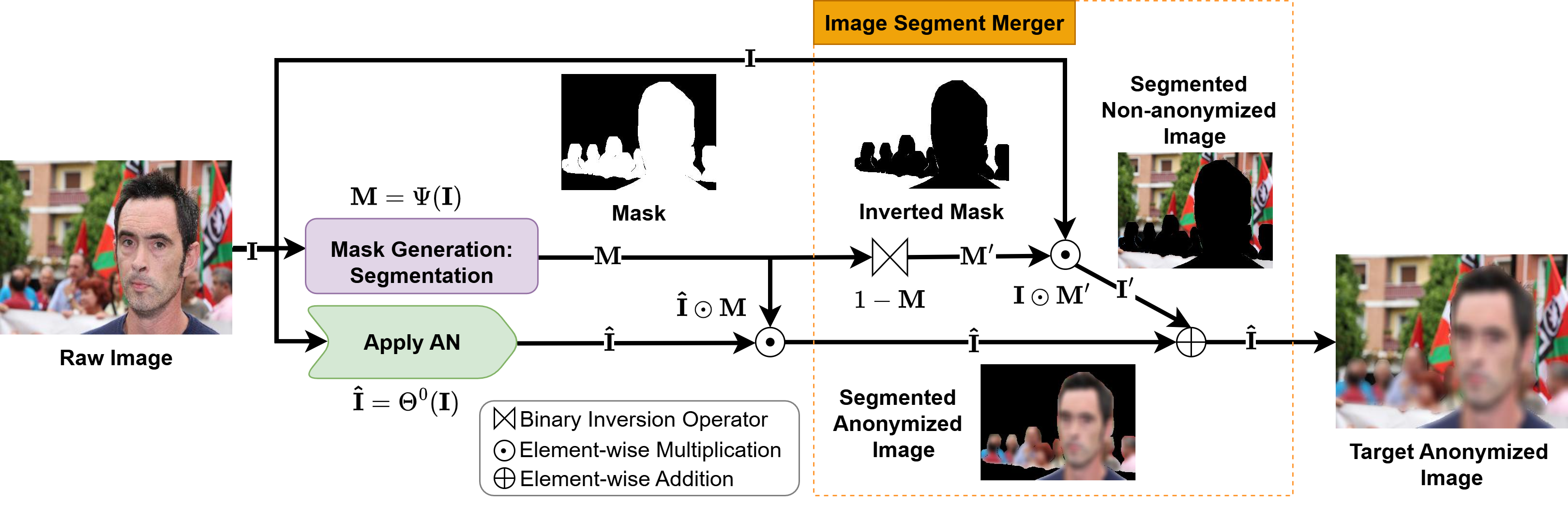} 
\caption{}
\label{fig:an_pipe_diags_nonadaptive}
\end{subfigure}
\begin{subfigure}[]{0.8\textwidth}
\centering
\includegraphics[width=1\textwidth]{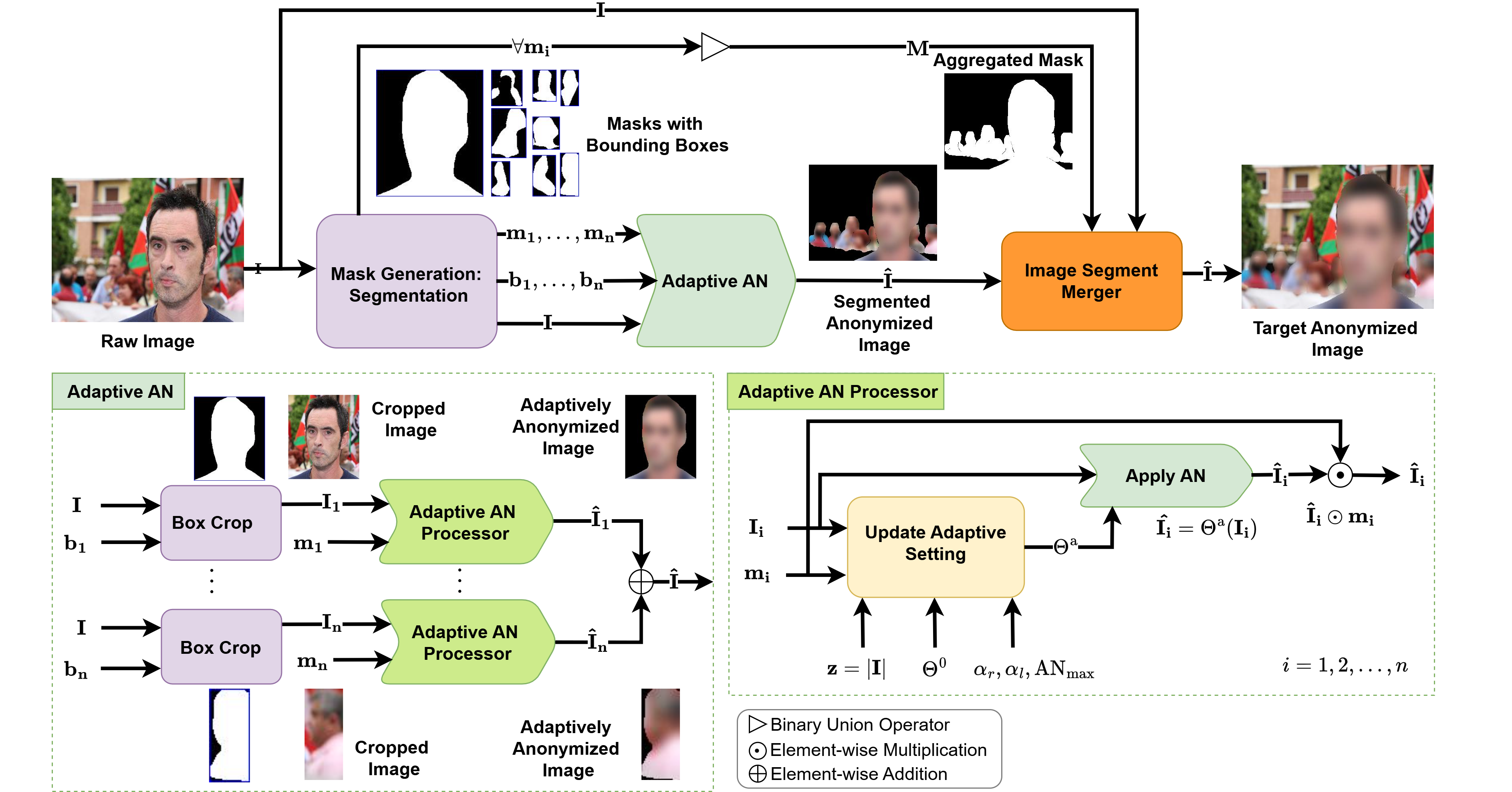}
\caption{}
\label{fig:an_pipe_diags_adaptive}
\end{subfigure}
\caption{System pipeline diagram of the AN systems: a) non-adaptive AN ($\mathcal{F}_\Theta^\text{0}$) that applies the same AN to all the target person subjects in the image (see Eq.~\eqref{eq:an_non_adaptive}), and b) adaptive AN ($\mathcal{F}_\Theta^\text{a}$) that adjust the AN strength depending on surface area of the targets (see Eq.~\eqref{eq:an_adaptive}). We have employed $\theta=\mathcal{G}$ for generating the demonstration example images.}
\label{fig:an_pipe_diags}
\vspace*{-0.5\baselineskip}
\end{figure*}

The previous full-body AN studies employ a single semantically generated binary $\mathbf{M} \in \mathbb{B}^{[H \times W]}$ per image frame $\mathbf{I} \in \mathbb{Z}^{[H \times W]}$~\cite{hukkelaas2019deepprivacy, wu2020privacy, dave2022spact, fioresi2023ted}, and applied $\Theta$ without considering the varying transformation effects of $\Theta$ on the target objects in $\mathbf{I}$ (see Fig.~\ref{fig:an_pipe_diags_nonadaptive}). This leads to poor performance in $U$ or $Q$ due to the non-optimal AN.
In our study, we employ instantly segmented masks $\mathbf{M}={\mathbf{m_0}, \mathbf{m_1},\dots,\mathbf{m_{n-1}}}$ for each target (see Algorithm~\ref{alg:object_detection}), where $n$ is the number of masks, and the AN optimizes the $\Theta$ for each target mask $\mathbf{m} \in \mathbf{M}$ (see Fig.~\ref{fig:an_pipe_diags_adaptive}).
We employ the You Only Look Once (\textsc{YOLO}) model~\cite{redmon2016you} to generate the masks of the objects of interest (line 4 in Algorithm~\ref{alg:object_detection}). The \textsc{YOLO} incorporates state-of-the-art and computationally efficient models that were trained on the MS-COCO dataset, and we utilize it for its promising small sizes and fast inference speed. 
The \textsc{YOLO} includes several model versions with varying complexity and accuracy. Although the larger models provide more accurate results, we chose \textsc{YOLOv8m-seg} for our study to strike a balance between accuracy and computational speed. A related study in Ref.~\cite{lee2024balancing} has similarly utilized \textsc{YOLOv8m-seg} for face and vehicle license plate AN for an object detection application.

\begin{algorithm}
\caption{$\boldsymbol{\Psi}$: Object of Interest Detection}
\label{alg:object_detection}
\begin{algorithmic}[1]
\small
\State $\boldsymbol{\Psi}$ ({$\mathbf{I}, \lambda=0.25$})
\newline
Instant object, e.g., a person, segmentation using YOLO
\newline
\Comment $\mathbf{I} \in \mathbb{Z}^{[H \times W]}$ is the input image
\newline
\Comment $\lambda$ is the segmentation detection confidence threshold
\newline
\State $\mathbf{z} \gets | \mathbf{I} |$ \Comment get image height and width $\mathbf{z}=[h, w]$
\State $\bar{\mathbf{I}} \gets ImageResize(\mathbf{I}, [320, 240])$ \Comment resize image to $[320, 240]$ for enhanced segmentation
\State $\bar{\mathbf{M}} \gets YOLOSeg(\bar{\mathbf{I}}, \lambda)$ \Comment returns list of masks
\State $\mathbf{M} \gets ImageSizeRestore(\bar{\mathbf{M}}, \mathbf{z})$ \Comment resize masks to match the original image dimensions
\State $\mathbf{B} \gets EstimateBoundingBox(\mathbf{M})$ \Comment generate bounding boxes surrounding $\mathbf{M}$
\newline
\Return $\mathbf{M},\mathbf{B}$
\end{algorithmic}
\end{algorithm}

Our proposed \textsc{LA3D} employs adaptive full-body de-identification system $\mathcal{F}^\text{a}_\Theta$ that enhances privacy protection (see Algorithm~\ref{alg:adaptive_anony}). 
The $\mathcal{F}_{\Theta}^{\text{a}} (\mathbf{X})$ applies adaptive AN for each detected subject mask $\mathbf{m_i} \in \mathbf{M}$ (lines 4--10 in Algorithm~\ref{alg:adaptive_anony}).
\begin{equation}
\label{eq:an_adaptive}
    \mathbf{\hat{X}}: \mathbf{I} \odot (1 - \mathbf{M}) + \sum_{\mathbf{m_i} \in \mathbf{M}}\mathbf{m_i} \odot \Theta^{\text{a}} (\mathbf{I[m_i]}),~
    \text{~for}~\forall \mathbf{I} \in \mathbf{X}.
\end{equation}

\begin{algorithm}[]
\caption{\textsc{LA3D}: Our Adaptive Full-Body Anonymization}
\label{alg:adaptive_anony}
\begin{algorithmic}[1]
\small
\State $\boldsymbol{\mathcal{F}^\text{a}}$ ({$\mathbf{I}, \Theta^0, \alpha_r, \alpha_l, \text{AN}_\text{max}=0$})
\newline
Our adaptive approach dynamically optimizes parameters of blurring and pixelization for varying sizes/depths of objects of interest.
\newline
\Comment $\mathbf{I} \in \mathbb{Z}^{[H \times W]}$ is the input image or video frame
\newline
\Comment $\Theta^0 \in \{\mathcal{G}, \mathcal{P}\}$ is the selected base AN function 
\newline
\Comment $\alpha_r$ is a hyperparameter for a global adaptive scaling
\newline
\Comment $\alpha_l$ is a hyperparameter for the upper limit for adaptive AN
\newline
\Comment $\text{AN}_\text{max}$ is a hyperparameter to enforce maximum adaptive AN 
\newline
\State $\mathbf{\hat{I}} \gets  \mathbf{I}[:]$ \Comment copy $\mathbf{I}$ into $\mathbf{\hat{I}}$
\State $\mathbf{M},\mathbf{B} \gets \Psi({\mathbf{I}})$ \Comment target masks $\mathbf{M}$ and bounding boxes $\mathbf{B}$
\For{$\mathbf{m_i} \in \mathbf{M}, \mathbf{b_i}\in \mathbf{B}$}  
    \Comment for each mask $\mathbf{m_i}$ and its box $\mathbf{b_i}$
    \State $r \gets \Omega (\mathbf{I}, \mathbf{m_i},  \alpha_r)$  \Comment adaptive scaling factor using Eq.~\eqref{eq:adaptive_r}
    \State ${\mathbf{I_i}} \gets {\mathbf{I}}[\mathbf{b_i} ]$ \Comment box-cropped image for a given $\mathbf{m_i}$ 
    \If {$\Theta^0 \in \mathcal{G}$}
        \State $\mathbf{\hat{I}}[\mathbf{m_i}] \gets \mathcal{G}^\text{a}({\mathbf{I_i}}, \Theta^0, r, \alpha_l, \text{AN}_\text{max})[\mathbf{m_i}]$ \Comment apply Alg.~\ref{alg:adaptive_anony_blur}
    \ElsIf {$\Theta^0 \in \mathcal{P}$} 
        \State $\mathbf{\hat{I}}[\mathbf{m_i}] \gets \mathcal{P}^\text{a}({\mathbf{I_i}}, \Theta^0, r, \alpha_l, \text{AN}_\text{max})[\mathbf{m_i}]$ \Comment apply Alg.~\ref{alg:adaptive_anony_pixelize}
    \EndIf
\EndFor
\newline
\Return $\mathbf{\hat{I}}$
\end{algorithmic}
\end{algorithm}

\vspace*{-\baselineskip}

The adaptive mechanism of the \textsc{LA3D} scales the AN strength using $\Omega$, depending on the relative area of the mask to the input image, as (line 5 in Algorithm~\ref{alg:adaptive_anony}): 
\begin{equation}
\label{eq:adaptive_r}
\begin{split}
   r_i & = \Omega (\mathbf{I}, \mathbf{m_i},  \alpha_r) = \max\left\{ \alpha_r \ln \left( \frac{100 \parallel \mathbf{m_i} \parallel}{\parallel \mathbf{I} \parallel}\right), 1 \right\}, \\
& \parallel \mathbf{I} \parallel = HW,~\parallel \mathbf{m_i} \parallel=\sum_{p_j \in \mathbf{m_i}}p_j,~\text{for}~\forall\mathbf{m_i} \in \mathbf{M},
\end{split}
\end{equation}
{where $\mathbf{I}$ is the input image, and $\mathbf{m_i} \in \mathbf{M}$ is the $i^\text{th}$ segmented binary mask. The $\parallel \parallel$ is a function that calculates surface area as the product of height and width for the $\mathbf{I}$, and as the sum of pixels $p_j \in \{0,1\}$ for the irregularly shaped mask $\mathbf{m_i}$. The adaptive AN for each mask $\mathbf{m_i}$ is formulated as: }
\begin{equation}
\label{eq:adaptive_Im}
\begin{split}
\mathbf{\hat{I}[m_i]} & = \Theta^{\text{a}} (\mathbf{I[m_i]}) : \Theta(\mathbf{I[m_i]}, r_i\langle\Theta^0\rangle),~\text{for}~\forall\mathbf{m_i} \in \mathbf{M},
\end{split}
\end{equation}
where $\langle \rangle$ denotes a parameter retrieving function, and the superscripts $0$ and $a$ indicate correspondence to the base (non-adaptive) and adaptive AN, respectively. The parameter $r_i$ is the AN adaptive factor of $\Theta^{\text{a}}$, and its value ranges $r \in [1, \alpha_r \ln (100)]$, as $\parallel \mathbf{m_i} \parallel \leq \parallel \mathbf{I} \parallel$. The $r_i$ value is calculated as a log of the relative area of the mask scaled by a hyperparameter $\alpha_r$. The log function provides a damping effect that enables smoother transitions among varying sizes of $\mathbf{m_i}$ as the parameter $r$ adjusts the AN strength of the base $\Theta^0$. 
We incorporate the hyperparameter $\alpha_r$ into the adaptive scaling to handle the size variation of the input image $\mathbf{I}$ (line 5). The parameter $\alpha_r$ enables consistent AN across different image scales of $\mathbf{I}$, and its value can be heuristically assigned or approximated as $\alpha_r = \mathbf{z}/\mathbf{z_\text{ref}}$, where $\mathbf{z}$ is the size of $\mathbf{I}$, and $\mathbf{z_\text{ref}}$ is a reference size for a unit scaling. We recommend generally tuning $\alpha_r$ or $\mathbf{z_\text{ref}}$, depending on the selected $\Theta^0$ method, to scale the AN with the image resolution.

The scaling mechanism of the adaptive AN also depends on the employed $\Theta^{0}$: for $\mathcal{G}$ at lines 7--8 and $\mathcal{P}$ at lines 9--10 in Algorithm~\ref{alg:adaptive_anony}. The algorithm adjusts value of the $\mathbf{k}$ and the $\mathbf{d}$ parameters for $\mathcal{G}$ in Algorithm~\ref{alg:adaptive_anony_blur} and $\mathcal{P}$ in Algorithm~\ref{alg:adaptive_anony_pixelize}, respectively. 
We also include a couple of global hyperparameters to the AN, i.e., a bound scaler $\alpha_l$ and a maximum AN flag $\text{AN}_\text{max}$, to further leverage flexibility to the adaptive AN. 
The $\alpha_l \in (0, 1]$ enforces a upper bound to the adaptive $\Theta^{\text{a}}$ ($\mathcal{G}$: line 6 in Algorithm~\ref{alg:adaptive_anony_blur} and $\mathcal{P}$: line 6 in Algorithm~\ref{alg:adaptive_anony_pixelize}). The boundaries are given as:
\begin{equation}
\label{eq:adaptive_boundary}
\begin{split}
    \mathbf{k^a_i} & = r_i\mathbf{k^0},~\text{for}~\mathbf{k^a_i}~\in~[\mathbf{k^0}, \alpha_l\mathbf{z_i}],~\alpha_l \in (0, 1], \\
    \mathbf{d^a_i} & = r_i\mathbf{d^0},~\text{for}~\mathbf{d^a_i}~\in~[\mathbf{d^0}, \alpha_l\mathbf{z_i}],~\alpha_l \in (0, 1],
    \end{split}
\end{equation}
where the $\mathbf{z_i}=[H_i, W_i]$ is the size of the rectangular bounding box of the mask $\mathbf{m_i}$. 
The $\text{AN}_\text{max} \in \{0, 1\}$ allows choosing between the two main adaptive AN mechanisms:
\begin{itemize}
    \item \textit{Bounded adaptive AN} ($\text{AN}_\text{max}=0$): implements bounded AN using the parameters of base $\Theta^0$, according to Eq.~\eqref{eq:adaptive_r} and Eq.~\eqref{eq:adaptive_boundary}. See lines 3--15 in Algorithm~\ref{alg:adaptive_anony_blur} for $\mathcal{G}$, and lines 2--6 in Algorithm~\ref{alg:adaptive_anony_pixelize} for $\mathcal{P}$. 
    \item \textit{Maximum adaptive AN} ($\text{AN}_\text{max}=1$): applies the maximum AN to all masks, ignoring the parameters of $\Theta^0$ and the hyperparameters. See lines 16--21 in Algorithm~\ref{alg:adaptive_anony_blur} for $\mathcal{G}$, and for $\mathcal{P}$: lines 7--8 in Algorithm~\ref{alg:adaptive_anony_pixelize} for $\mathcal{P}$. For example, the utmost AN for the $\mathcal{P}$ results in a single color silhouette for $ \mathbf{d^a_i}=\mathbf{z_i}$, as the entire region of $\mathbf{m_i}$ will be downsampled into a single pixel before the upsampling.  
\end{itemize}

Furthermore, the bounded adaptive AN with $\mathcal{G}$ has a flag $\text{G}_\text{full}$ that the adaptive scaling applies to both of $\mathbf{k}$ and $\boldsymbol{\sigma}$ (lines 10--11 in Algorithm~\ref{alg:adaptive_anony_blur}) if set else only to the $\mathbf{k}$ (lines 12--13).

\begin{algorithm}[]
\caption{$\boldsymbol{\mathcal{G}^\text{a}}$: Adaptive Gaussian Blurring}
\label{alg:adaptive_anony_blur}
\begin{algorithmic}[1]
\small
\State $\boldsymbol{\mathcal{G}^\text{a}}$ ({$\mathbf{I_i}, \mathcal{G}^0, r, \alpha_l, \text{AN}_\text{max}=0, \text{G}_\text{full}=0$})
\newline
Our adaptive Gaussian blurring algorithm.
\newline
\Comment $\mathbf{I_i} \in \mathbb{Z}^{[H_i \times W_i]}$ is a box-cropped image for a given mask
\newline
\Comment $\mathcal{G}^0$ is a Gaussian blurring function with base parameters
\newline
\Comment $r$ is the adaptive scaling factor based on relative mask size
\newline
\Comment $\text{G}_\text{full}$ is a hyperparameter to include scaling the kernel's $\sigma$ 
\newline
\State $\mathbf{z_i} \gets |\mathbf{I_i}|$ \Comment get image height and width $\mathbf{z_i}=[H_i, W_i]$

\If {$\text{AN}_\text{max}=0$} \Comment for maximum adaptive AN
 \State $\mathbf{k^0}, \boldsymbol{\sigma^0} \gets \langle\mathcal{G}^0\rangle$ \Comment get parameters of the base kernel
    \State $\mathbf{k^a_i} \gets r\mathbf{k^0}$ \Comment adaptive $\mathbf{k}$
    \State $\mathbf{k^a_i} \gets \text{min} \left\{\mathbf{k^a_i}, \text{max}\{\alpha_l\mathbf{z_i} , 1\} \right\}$ \Comment limit maximum $\mathbf{k^a_i}$
    \If {$\mathbf{k^a_i}\%2=0$} \Comment check even $\mathbf{k^a_i}$ via module $\%$ operator
        \State $\mathbf{k^a_i} \gets \mathbf{k^a_i} - 1$ \Comment make $\mathbf{k^a_i}$ an odd number
    \EndIf
    \If {$\text{G}_\text{full}=1$} \Comment check if to include $\boldsymbol{\sigma}$ in the adaptive AN
        \State $\boldsymbol{\sigma^a_i} \gets r\boldsymbol{\sigma^0}$  \Comment adaptive $\boldsymbol{\sigma}$
    \Else
        \State $\boldsymbol{\sigma^a_i} \gets \boldsymbol{\sigma^0} $  \Comment non-adaptive $\boldsymbol{\sigma}$
    \EndIf
     \State $\boldsymbol{\sigma^a_i} \gets \text{min} \left\{\boldsymbol{\sigma^0}, \mathbf{k^a_i} \right\}$ \Comment limit to keep the smoothing effect
\Else     
    \State $\mathbf{k^a_i} \gets \mathbf{z_i}$ \Comment set $\mathbf{k^a_i}$ to a maximum boundary value for AN
    \If {$\mathbf{k^a_i}\%2=0$} \Comment check even $\mathbf{k^a_i}$ via module $\%$ operator
        \State $\mathbf{k^a_i} \gets \mathbf{k^a_i} - 1$ \Comment make $\mathbf{k^a_i}$ an odd number
    \EndIf
    \State $\boldsymbol{\sigma^a_i} \gets 0.3(0.5(\mathbf{k^a_i}-1)-1) + 0.8$ \Comment adopted from OpenCV
 \EndIf
\State $\mathbf{\hat{I}_i} \gets \mathcal{G}_{\mathbf{k^a_i}, \boldsymbol{\sigma^a_i}}(\mathbf{I_i})$ \Comment blurred image using $\mathbf{k^a_i}$ and $\boldsymbol{\sigma^a_i}$, Eq.~\eqref{eq:gaussian_kernel}
\newline
\Return $\mathbf{\hat{I}_i}$
\end{algorithmic}
\end{algorithm}

\begin{algorithm}[]
\caption{$\boldsymbol{\mathcal{P}^\text{a}}$: Adaptive Pixelization}
\label{alg:adaptive_anony_pixelize}
\begin{algorithmic}[1]
\small
\State $\boldsymbol{\mathcal{P}^\text{a}}$ ({$\mathbf{I_i}, \mathcal{P}^0, r, \alpha_l, \text{AN}_\text{max}=0$})
\newline
Our adaptive pixelization algorithm.
\newline
\Comment $\mathbf{I_i} \in \mathbb{Z}^{[H_i \times W_i]}$ is a box-cropped image for a given mask
\newline
\Comment $\mathcal{P}^0$ is the pixelization function with base parameters
\newline
\Comment $r$ is the adaptive scaling factor based on relative mask size
\newline
\If {$\text{AN}_\text{max}=0$} \Comment for maximum adaptive AN
    \State $\mathbf{d^0} \gets \langle\mathcal{P}^0\rangle$ \Comment get base downsizing factor
    \State $\mathbf{z_i} \gets | \mathbf{I_i} |$ \Comment get image height and width $\mathbf{z_i}=[H_i, W_i]$
    \State $\mathbf{d^a_i} \gets r\mathbf{d^0}$ \Comment adaptive $\mathbf{d}$
    \State $\mathbf{d^a_i} \gets \text{min} \left\{\mathbf{d^a_i}, \text{max}\{\alpha_l\mathbf{z_i}, 1\} \right\}$ \Comment limit maximum $\mathbf{d^a_i}$
\Else
     \State $\mathbf{d^a_i} \gets \mathbf{z_i} $ \Comment set $\mathbf{d^a_i}$ to a maximum boundary value for AN
\EndIf
\State $\mathbf{\hat{I}_i} \gets \mathcal{S}_{\text{up}}(\mathcal{S}_{\text{down}}(\mathbf{I_i}, \mathbf{d^a_i}), \mathbf{z_i})$ \Comment downsampling $\mathbf{I_i}$ by $\mathbf{d^a_i}$ and restoring size to $\mathbf{z_i}$ by interpolation to the nearest pixel, Eq.~\eqref{eq:pixelize_fn}
\newline
\Return $\mathbf{\hat{I}_i}$
\end{algorithmic}
\end{algorithm}

\subsection{Privacy Attribute Classification Model}

\begin{figure}[htbp]
\centering
\includegraphics[width=0.4\columnwidth]{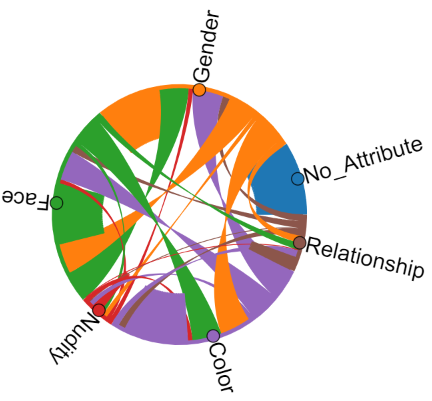} 
\caption{The privacy attribute class distribution of the VISPR train dataset~\cite{orekondy2017towards}, showing class imbalance and co-occurrence.}
\label{fig:pd_vispr_multi__train_labels}
\vspace*{-\baselineskip}
\end{figure}

The ability to extract biometric data, such as face, gender, and ethnicity attributes, constitutes concerns for privacy~\cite{meden2021privacy}. 
We thus employ a multi-label classifier \textsc{ResNet50}~\cite{he2016deep} model, $\mathcal{F}_P$, for privacy attribute leakage detection (PD). 
We select six widely utilized privacy class attributes from the VISPR dataset~\cite{orekondy2017towards}, following~\cite{wu2020privacy, dave2022spact, fioresi2023ted} (as described in Table~\ref{tbl:pd_vispr_dataset_classes}).
The $ \mathcal{F}_P$ model is formulated as: 
\begin{equation}
    \mathcal{F}_P: \textsc{ResNet50}_{\text{e}}(N_f) \rightarrow \textsc{FC}(N_c) \rightarrow \mathcal{A}(N_c),
\end{equation}
where $\mathbf{c}=\mathcal{F}_P(\mathbf{I})$ is the set of predicted PD class labels of $\mathbf{I}$ with $\mathbf{c}=\{c_1, \dots, c_{N_c}\}$, $c_i \in \{0,1\}$, and the $N_c=6$ is the number of classes. The $\textsc{ResNet50}_{\text{e}}$ is the image encoding network that extracts $\mathbf{I_e} \in \mathbb{R}^{[1 \times N_f]}$ feature map from the input image $\mathbf{I} \in \mathbb{Z}^{[H \times W]}$, and the $\textsc{FC}$ is the final classification network with {sigmoid} ($\mathcal{A}$) activation function. 
We initialize the model with a pre-trained (on ImageNet) \textsc{ResNet50} image encoder~\cite{he2016deep, he2015delving} and then fine-tune it on the VISPR dataset with a class-weighted loss function $\mathcal{L}_w$.
\begin{equation}
    \begin{aligned}
    \label{eq:weighted_loss_fn}
    \mathcal{L}_w = \sum_{i=1}^{N_c} W_{i}\mathcal{L}_{i},~\text{where}~W_{i}=1/D_{i},~D_{i}=n_{i}/n_c,
    \end{aligned}
\end{equation}
where the $\mathcal{L}_{i}$ is a {binary cross entropy loss} function, the $W_{i}$ is a weighting factor, and the $D_{i}$ is class distribution for the $i^\text{th}$ class label. The $n_{i}$ and $n_c$ are the number of samples with active label for class $i$ and the total number of samples in the dataset, respectively.
The class label distribution of the training-set of the VISPR, shown in Fig.~\ref{fig:pd_vispr_multi__train_labels}, indicates that the {Nudity} and {Relationship} classes have relatively low contributions. The $\mathcal{F}_P$ would struggle to learn these classes effectively due to the class label imbalance and label co-occurrence with the other dominant class labels.   
The $\mathcal{L}_{i}$ is calculated per target class label and summed after scaling with the class weight $W_{i}$ to estimate the final loss score $\mathcal{L}_w$. 
We estimate $W_i = 1/D_i$ to compensate for the imbalance and improve the classification performance of $\mathcal{F}_P$. 
We utilize a learning rate of $10^{-3}$ with a linear warmup with a step-based scheduler that reduces the rate by $1/5$ when the training loss saturates. We trained the model up to 100 epochs using \textsc{Adam} optimizer with L2 regularization of $10^{-5}$.

\begin{table}[] \small
    \centering
    \caption{\textsc{Privacy Attribute Classes of the VISPR Dataset~\cite{orekondy2017towards}.}}
    \resizebox{\columnwidth}{!}{
    \begin{tabular}{llll}
    \hline
    Class ID & Class Name & Description    &    Training-set $D$             \\ \hline
    1                 & No\_Attribute       & No attributes revealing privacy & 51.53\%\\ 
    2                 & Gender              & Person gender          & 46.25\%                  \\ 
    3                 & Face                & Full or partial face  & 44.18\%                   \\ 
    4                 & Nudity              & Partial nudity       & 5.88\%                    \\ 
    5                 & Color               & Skin color        & 47.10\%                       \\ 
    6                 & Relationship        & Personal and social relationship   & 9.64\%      \\ \hline
    \end{tabular}
    }
    \noindent\footnotesize{\par  
The $D$ is a multi-label distribution percentage.
\par}
\label{tbl:pd_vispr_dataset_classes}
\vspace*{-0.5\baselineskip}
\end{table}

\subsection{Video Anomaly Detection Model}

We employ state-of-the-art WSAD models, such as \textsc{PEL4VAD}~\cite{pu2024learning} and \textsc{MGFN}~\cite{chen2023mgfn}, to detect anomalies in video datasets after AN. 
These models are among the top-performing methods in VAD~\cite{chen2023mgfn, sharif2023cnn, pu2024learning}. 

The VAD models operate on extracted features from a sequence of frames using pre-trained video encoders~\cite{sharif2023cnn, carreira2017quo, wang2018non, zhu2020comprehensive}. 
The video encoder $\mathcal{F}_\text{e}$ generates embedded features for given a video data $\mathbf{X} \in \mathbb{R}^{N \times H \times W \times 3}$, where $N, H$ and $W$ denote the number of frames, height, and width of $\mathbf{X}$, respectively. The video is evenly segmented, $V$ frames per segment, into $T$ clips. The $\mathcal{F}_\text{e}$ provides feature maps as $\mathcal{F}_\text{e}: \mathbf{X} \rightarrow \mathbf{X}_{e} \in \mathbb{R}^{T \times B \times D}$, where $B$ is the number of crop augmentations, and $D$ is the feature dimension. 
Several pre-trained video feature encoders are available in the literature~\cite{sharif2023cnn}, and most employ C3D~\cite{ji20123d, sultani2018real, sharif2023cnn} and I3D~\cite{carreira2017quo, sharif2023cnn, pu2024learning, chen2023mgfn} methods. Recent studies demonstrate the superiority of I3D in diverse CV applications~\cite{dave2022spact, sharif2023cnn, chen2023mgfn, fioresi2023ted, pu2024learning}. 
Thus, we utilize a pre-trained \textsc{ResNet50-I3D} video encoder~\cite{carreira2017quo, wang2018non} 
that was trained on the Kinetics400 action dataset~\cite{kay2017kinetics}.
We refer readers to Ref.~\cite{zhu2020comprehensive} for further review on video encoders.

\section{Results and Discussion}
\label{sec:results_and_discussion}

This section presents the results of our AN study on several public datasets, including VISPR, Market1501, UCF-Crime, and XD-Violence. 
We follow the standard training and test split for all datasets. 
We implement and compare the widely utilized privacy-protection techniques, presented in Table~\ref{tbl:anony_methods}, along with our approaches. 
To maintain a fair comparison across methods, we apply the baseline settings utilized in recent benchmark studies of DL-based AN models~\cite{dave2022spact, fioresi2023ted, wu2020privacy, hukkelaas2019deepprivacy, hukkelaas2023deepprivacy2}. 
We will compare the performance of our approach with recent DL-based AN methods, and their configuration summary is provided in Table~\ref{tbl:dl_benchmark_models} in the Appendix~\ref{sec:results_dl_benchmark}.

\begin{table}[] 
\caption{Anonymization Methods and their Configurations.}
\resizebox{1\columnwidth}{!}{
\begin{tabular}{lp{7.0cm}}
\hline
{Method}         & {Description} \\ \hline
{No-AN}              &    The raw footage without AN.        \\ 
{MASKED}              &   AN by black silhouette~\cite{dave2022spact, fioresi2023ted, wu2020privacy, hukkelaas2019deepprivacy, hukkelaas2023deepprivacy2}. \\ 
{EDGED}                  &   AN by Canny edge detection with intensity gradient ($\rho$) thresholds $\upsilon_\text{min}=100$ and $\upsilon_\text{max}=200$~\cite{canny1986computational}; $\rho>\upsilon_\text{max}$: edges, $\rho<\upsilon_\text{min}$: non-edges, and otherwise: decide based on connectivity~\cite{canny1986computational}.  \\ 
{$\mathcal{G}^\text{0}$}                  &     $\mathcal{G}^\text{0}$: AN by $\mathcal{G}_{\mathbf{k},\boldsymbol{\sigma}}$  with a $\mathbf{k}=[13,13]$ and $\boldsymbol{\sigma}=[10,10]$, following Refs.~\cite{dave2022spact, fioresi2023ted}.   \\ 
{$\mathcal{G}^\text{a}(\alpha_l=\lambda)$}        &   $\mathcal{G}^\text{a}$: $\mathcal{G}^\text{0}$ leveraged by our adaptive AN with $\alpha_r=1$ (unless specified), and $\alpha_l=\lambda$.  \\ 
{$\mathcal{G}^\text{a} _\text{max}$}        &   $\mathcal{G}^\text{a}$: $\mathcal{G}^\text{0}$ leveraged by our adaptive AN with maximum AN setting of $\mathbf{k^a_i} = \mathbf{z_i}$ with $\text{AN}_\text{max}=1$.  \\ 
{$\mathcal{P}^\text{0}_{d \in \{2,4,8\}}$}             &  $\mathcal{P}^\text{0}$: AN by $\mathcal{P}_{\mathbf{d}}$ that linearly downsamples the image by a factor of $\mathbf{d}=[d,d]$ and then interpolates back to the original size, following Refs.~\cite{dave2022spact, fioresi2023ted, wu2020privacy, hukkelaas2019deepprivacy, hukkelaas2023deepprivacy2}.    \\ 
{$\mathcal{P}^\text{a}_{d \in \{2,4,8\}}(\alpha_l=\lambda)$}      & $\mathcal{P}^\text{a}$: $\mathcal{P}^\text{0}_d$ leveraged by our adaptive AN with $\alpha_r=1$ (unless specified), and $\alpha_l=\lambda$. 
\\ \hline
\end{tabular}
}
\label{tbl:anony_methods}
\vspace*{-0.5\baselineskip}
\end{table}

\subsection{Evaluation Approach and Performance Metrics}

Visual AN techniques can be evaluated in different ways to assess their efficacy in privacy protection and the retention of non-sensitive utilities after the de-identification~\cite{zhang2024understanding, dave2022spact, fioresi2023ted, hukkelaas2023deepprivacy2}. 
We will focus on recognition attack mechanisms to measure privacy protection via: 1) person attribute detection (PD) metrics that measure the leakage on the multi-label privacy attributes~\cite{dave2022spact, fioresi2023ted}, and 2) person re-identification (ReID) metrics that measure the recognition rate of a person from body image~\cite{hukkelaas2023deepprivacy2, hukkelaas2023realistic}. 
We will evaluate the utility retention quality of the anonymized data through performance in a VAD~\cite{fioresi2023ted}. 
Due to the lack of adequate public datasets that incorporate both privacy and VAD, we will follow protocols of cross-dataset evaluation to measure the efficacy of the AN methods~\cite{dave2022spact, fioresi2023ted, wu2020privacy, hukkelaas2019deepprivacy, hukkelaas2023deepprivacy2, hukkelaas2023realistic}. 
Alongside quantitative assessment using CV tools, AN can also be evaluated qualitatively using human visual inspection. While this study does not include a comprehensive human vision analysis, we will provide several images to illustrate the use cases of AN.

{We will evaluate the PD through the multi-label classification $\mathcal{F}_P$ model using {average precision} (AP) and {recall} (R) on the VISPR dataset. The AP is the area under the {precision-recall curve}, and the R is calculated at a threshold of $0.5$ for each class label. We provide the aggregate performance using mean AP (mAP) and mean R (mR), which are the averages of the AP and the R over the class labels, respectively: }
\begin{equation}
\label{eq:metrics_map_mr}
\begin{aligned}
    \text{mAP} &= \frac{1}{N_c} \sum_{i=0}^{N_c-1}\text{AP}_i, ~~
    \text{mR} = \frac{1}{N_c} \sum_{i=0}^{N_c-1}\text{R}_i,
    \end{aligned}
\end{equation}
where $N_c$ is the number of classes, and $\text{AP}_i$ and $\text{R}_i$ are the AP and R of the $i^\text{th}$ class. Lower scores of the mAP and mR indicate better potency in anonymizing the privacy attributes. 

{We will employ a popular metric, {cumulative matching characteristics curve} (CMC), in addition to the mAP for the ReID evaluation on the Market1501 dataset. The CMC-Rk accuracy considers the correct match of the query identity from the top-k ranked gallery samples, based on predicted scores. The mAP is calculated using Eq.~\eqref{eq:metrics_map_mr}, where the classes represent the person identities. Lower scores of the CMC and mAP indicate enhanced protection against ReID. 
}

We will evaluate the performance of the utility VAD using AP and the {area under the receiver operating characteristic curve} (AUC) on the UCF and XD-Violence datasets. The AUC is calculated from the plot curve of the {true positive rate} against the {false positive rate}.  
{Higher AP and AUC scores represent higher robustness in preserving the data utility quality for accurate VAD. Thereafter, in the results tables and figures below, we will utilize arrows ($\uparrow$, $\downarrow$, and $\leftarrow$) to indicate towards the better scores, and {bold font} will indicate the best score. We will also employ the $\Delta_\text{N}$ and $\Delta_\text{0}$ to indicate the relative score with respect to the No-AN and the non-adaptive ANs (for the $\mathcal{G}$ and $\mathcal{P}$), respectively. }

\subsection{Evaluation on Privacy Leakage Detection}

\begin{figure*}[hb]
\centering
\setlength{\kw}{1.5cm}
\begin{subfigure}[]{0.75\textwidth}
\centering
\begin{tabular}{K{\kw}K{\kw}K{\kw}K{\kw}K{\kw}K{\kw}K{\kw}}
No-AN  & MASKED & EDGED & $\mathcal{P}^\text{0}_2$ & $\mathcal{P}^\text{0}_4$ & $\mathcal{P}^\text{0}_8$ & $\mathcal{G}^\text{0}$
\end{tabular}
\includegraphics[width=1\linewidth]{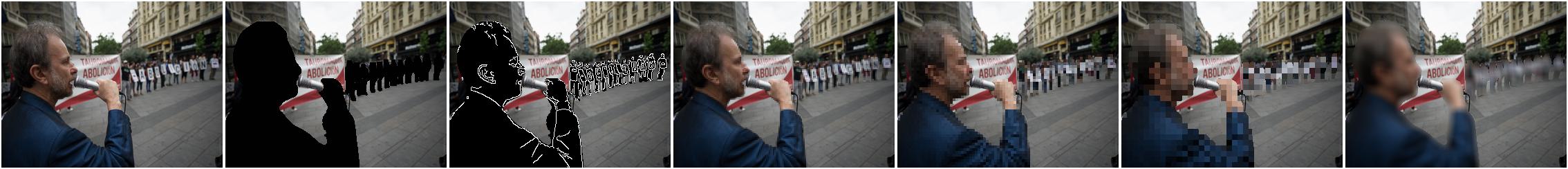} \\
\includegraphics[width=1\linewidth]{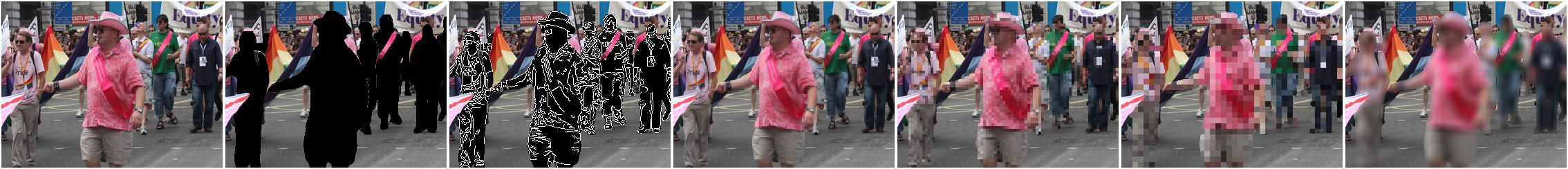}
\vspace{-0.1cm}
\end{subfigure}
\begin{subfigure}[]{0.64\textwidth}
\centering
\begin{tabular}{K{\kw}K{\kw}K{\kw}K{\kw}K{\kw}K{\kw}} 
$\mathcal{P}^\text{a}_2$ ($\alpha_r=1.0$, $\alpha_l=0.5$) & $\mathcal{P}^\text{a}_4$ ($\alpha_r=1.0$, $\alpha_l=0.5$)  & $\mathcal{P}^\text{a}_8$ ($\alpha_r=1.0$, $\alpha_l=0.5$) &  $\mathcal{P}^\text{a}_\text{max}$ ($\text{AN}_\text{max}=1$) & $\mathcal{G}^\text{a}$ ($\alpha_r=1.0$, $\alpha_l=0.5$) & $\mathcal{G}^\text{a}_\text{max}$ ($\text{AN}_\text{max}=1$)
\end{tabular}
\includegraphics[width=1\linewidth]{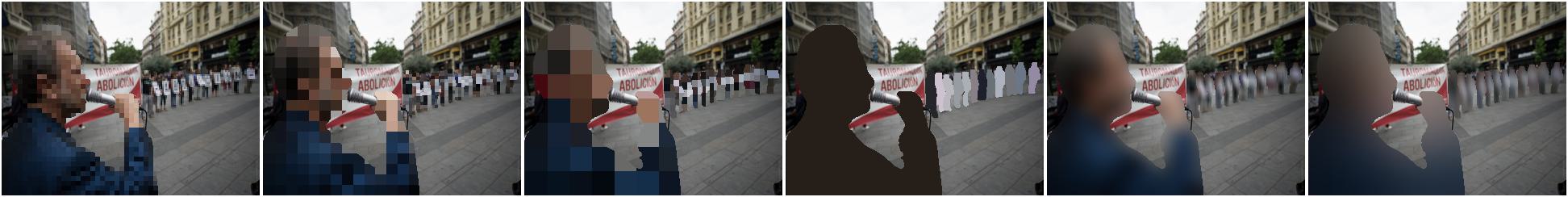} \\
\includegraphics[width=1\linewidth]{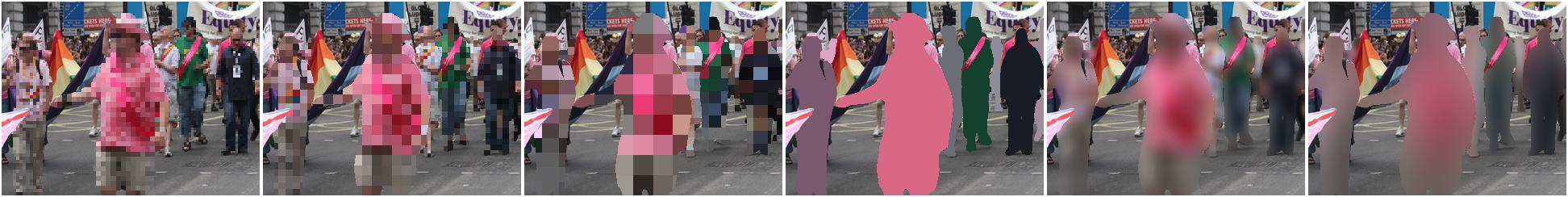}
\end{subfigure}
\caption{{Visual comparison of different AN methods on sample images from the VISPR dataset~\cite{orekondy2017towards}. The subscripts $0$ and $a$ denote non-adaptive and adaptive AN, respectively. The images have $\mathbf{z} = [320, 240]$ resolution.}
}
\label{fig:vispr_anony_compare_all_im_s320_240}
\vspace*{-0.5\baselineskip}
\end{figure*}

\begin{figure*}[thb]
\centering
\begin{subfigure}[]{1\textwidth}
\centering
\setlength{\kw}{1.5cm}
\resizebox{1\textwidth}{!}{
\begin{tabular}{p{1.7cm}K{\kw}K{\kw}K{\kw}K{\kw}K{\kw}K{\kw}K{\kw}K{\kw}}
& No-AN & $\mathcal{P}^\text{a}_4$ ($\alpha_r=1.0$, $\alpha_l=0.5$) & $\mathcal{P}^\text{a}_4$ ($\alpha_r=\mathbf{z}/\mathbf{z_\text{ref}}$, $\alpha_l=0.5$) & $\mathcal{P}^\text{a}_\text{max}$ ($\text{AN}_\text{max}=1$) & $\mathcal{G}^\text{a}$ ($\alpha_r=1.0$, $\alpha_l=0.5$) & $\mathcal{G}^\text{a}$ ($\alpha_r=\mathbf{z}/\mathbf{z_\text{ref}}$, $\alpha_l=0.5$) & $\mathcal{G}^\text{a}$ ($\alpha_r=\mathbf{z}/\mathbf{z_\text{ref}}$, $\alpha_l=0.5$, $\text{G}_\text{full}=1$) & $\mathcal{G}^\text{a}_\text{max}$ ($\text{AN}_\text{max}=1$)
\end{tabular}
}
\resizebox{1\textwidth}{!}{
\begin{tabular}{p{2.2cm}c}
\small $\mathbf{z} = [80, 60]$ & \includegraphics[width=1.0\linewidth]{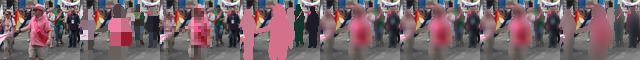} \\
\small $\mathbf{z} = [160, 120]$ & \includegraphics[width=1.0\linewidth]{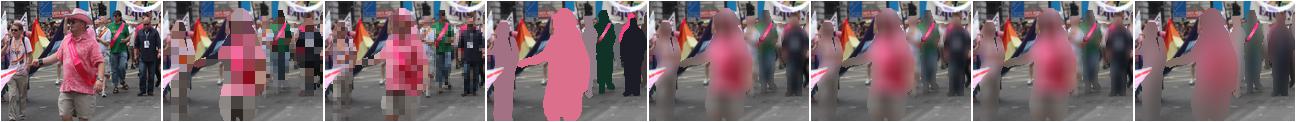} \\
\small $\mathbf{z} = [320, 240]$ & \includegraphics[width=1.0\linewidth]{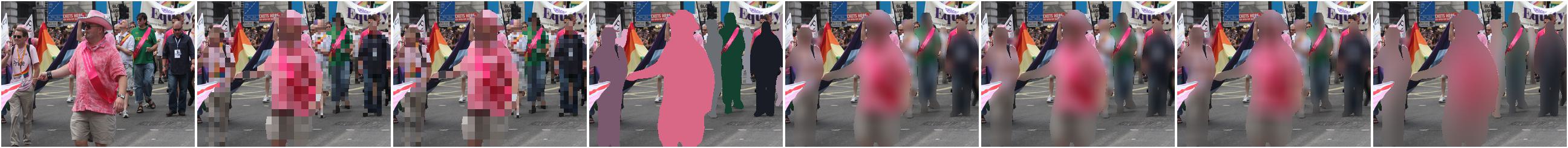} \\
\small $\mathbf{z} = [640, 480]$ & \includegraphics[width=1.0\linewidth]{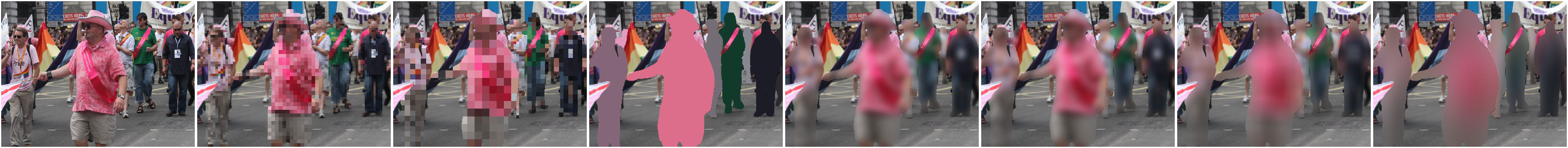} \\
\small $\mathbf{z} = [1280, 960]$ & \includegraphics[width=1.0\linewidth]{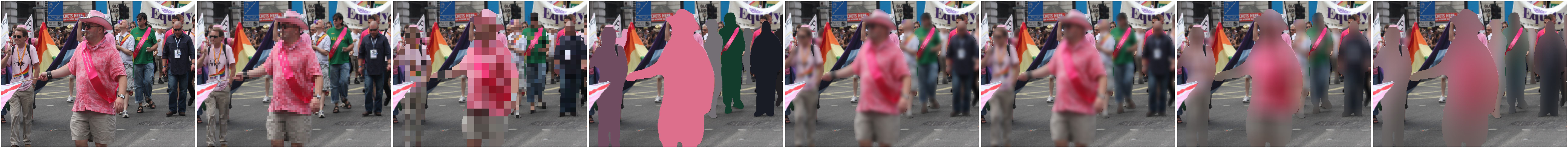}
\end{tabular}
}
\end{subfigure}
\caption{Adaptive AN and hyperparameter effect on images with different resolutions (top-down) from the VISPR dataset~\cite{orekondy2017towards}. The subscript $a$ denotes adaptive AN. The reference image size for estimating $\alpha_r$ is set to $\mathbf{z_\text{ref}}=[320, 240]$. The maximum adaptive AN with $\text{AN}_\text{max}=1$ is inherently scalable to image sizes, as $\mathbf{k^a_i} = \mathbf{z_i}$ and $\mathbf{d^a_i} = \mathbf{z_i}$. To maintain the bounded adaptive AN for $\text{AN}_\text{max}\neq 1$ as the reference resolution $\mathbf{z_\text{ref}}$, $\alpha_r =\mathbf{z}/\mathbf{z_\text{ref}}$ is necessary for the scaled image with $\mathbf{z} \neq \mathbf{z_\text{ref}}$.
}
\label{fig:vispr_anony_compare_all_im_scaled_compare_all}
\vspace*{-1\baselineskip}
\end{figure*}

\begin{figure}[htbp]
\centering
\setlength{\kw}{1.4cm}
\begin{subfigure}[]{1\columnwidth}
\centering
\begin{tabular}{K{\kw}K{\kw}K{\kw}K{\kw}K{\kw}}
No-AN & $\mathcal{P}^\text{0}_4$ & $\mathcal{P}^\text{a}_4$  & $\mathcal{G}^\text{0}$ & $\mathcal{G}^\text{a}$
\end{tabular}

\includegraphics[width=1\linewidth]{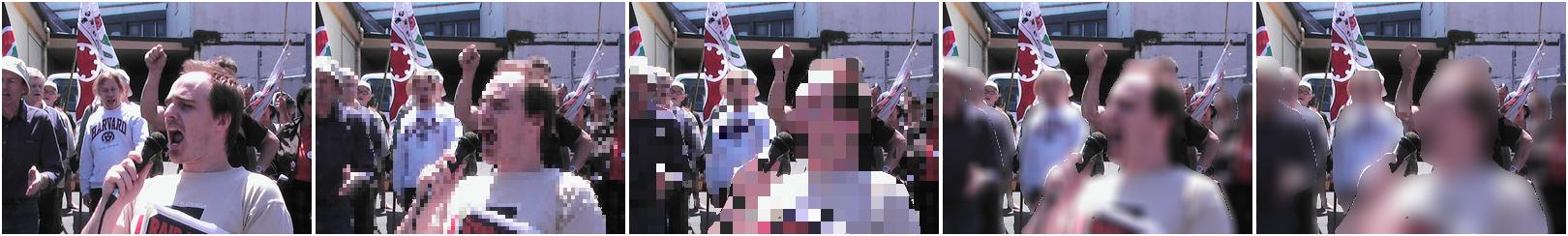} \\
\includegraphics[width=1\linewidth]{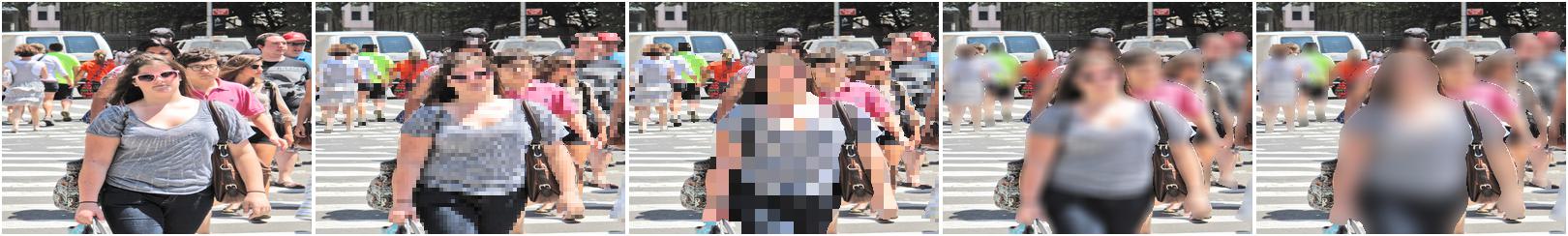} \\
\includegraphics[width=1\linewidth]{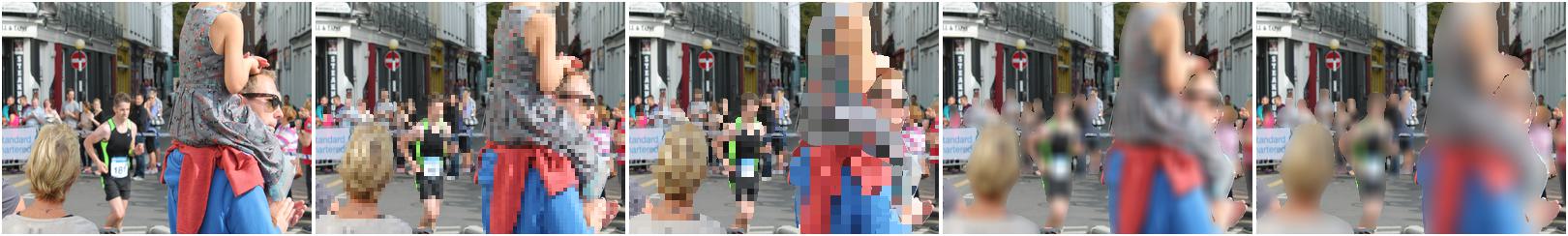} \\
\includegraphics[width=1\linewidth]{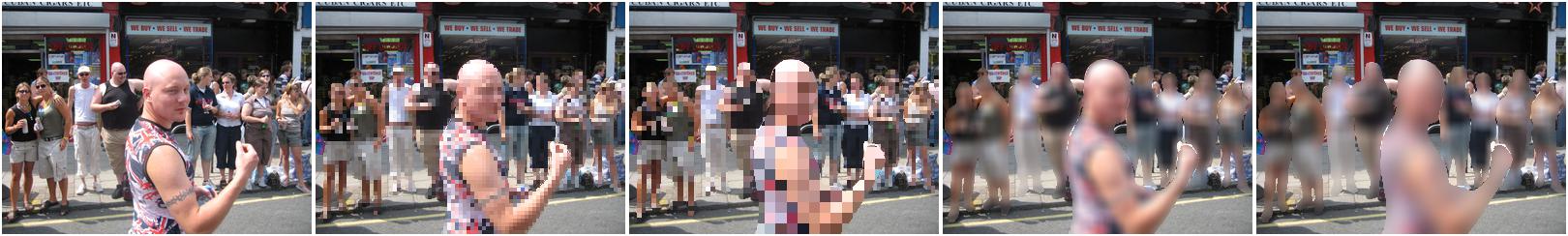} \\
\includegraphics[width=1\linewidth]{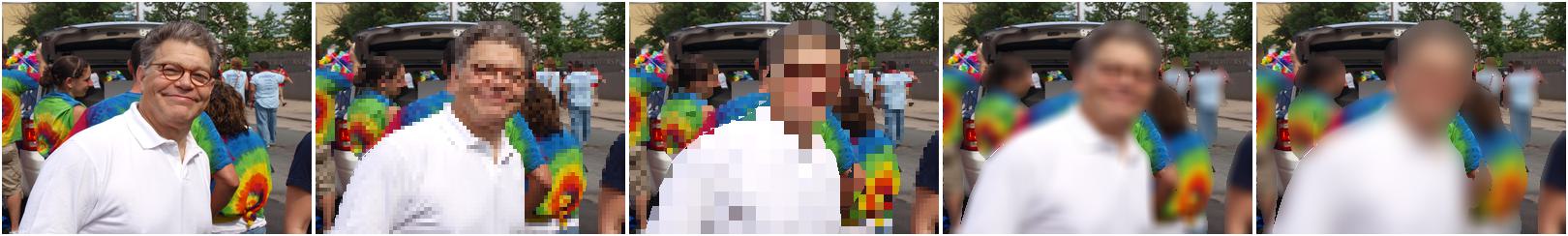} \\
\end{subfigure}
\caption{AN on person figures at varying depths on sample images (with resolution of $\mathbf{z} = [320, 240]$) from the VISPR dataset~\cite{orekondy2017towards}. The subscripts $0$ and $a$ denote non-adaptive and adaptive AN (with $\alpha_r=1.0$ and $\alpha_l=0.5$), respectively.
}
\label{fig:vispr_anony_compare_D4_adpative_im_s320_240_more}
\vspace*{-1\baselineskip}
\end{figure}

We will evaluate the privacy protection capability of the AN methods using PD following~\cite{dave2022spact, fioresi2023ted}, and ReID following~\cite{hukkelaas2023deepprivacy2}.
Before discussing the quantitative performance using CV models, we will first provide a qualitative comparison of the different AN methods and the impact of hyperparameters (as illustrated in Figs.~\ref{fig:vispr_anony_compare_all_im_s320_240}~and~\ref{fig:vispr_anony_compare_all_im_scaled_compare_all}). Fig.~\ref{fig:vispr_anony_compare_all_im_s320_240} illustrates several baseline and adaptive AN methods on sample images. The methods provide varying levels of AN, and the adaptive approaches have leveraged the AN compared to their corresponding baselines, as visually depicted. 
Non-adaptive $\mathcal{P}^\text{0}$ and $\mathcal{G}^\text{0}$, and lower adaptive $\mathcal{P}^\text{a}_2$ provide very limited protection and are ineffective for human vision inspections.
Fig.~\ref{fig:vispr_anony_compare_all_im_scaled_compare_all} demonstrates the scalability and impact of the AN hyperparameters (the $\alpha_l$ and $\alpha_r$) on varying image resolutions.  
The maximum adaptive ANs ($\text{AN}_\text{max}=1$) are inherently scalable, as $\mathbf{k^a_i} = \mathbf{z_i}$ and $\mathbf{d^a_i} = \mathbf{z_i}$. However, adjusting $\alpha_r = \mathbf{z}/\mathbf{z_\text{ref}}$, for the images with $\mathbf{z} \neq \mathbf{z_\text{ref}}$, is essential for the bounded ANs ($\text{AN}_\text{max}=0$) to maintain the protection as of the $\mathbf{z_\text{ref}}$ with $\alpha_r = 1$. For the blurring AN, the $\boldsymbol{\sigma}$ along with the $\mathbf{k}$ of the $\mathcal{G}$ need to be scaled, i.e., $\text{G}_\text{full}=1$, to keep the AN competency. 
Fig.~\ref{fig:vispr_anony_compare_D4_adpative_im_s320_240_more} illustrates further sample images to visually demonstrate the superiority of our adaptive AN over the baselines, where the images consist of person figures at different depths.

\subsubsection{Privacy Attribute Detection}
\label{sec:results_pd}

We evaluate PD to measure leakage on the multi-label privacy attributes, given in Table~\ref{tbl:pd_vispr_dataset_classes}, on the VISPR dataset~\cite{orekondy2017towards}.
The VISPR dataset comprises 22,000 publicly available Flickr images labeled with 68 privacy-related attributes.
The training, validation, and testing sets contain 10,000, 4,000, and 8,000 images, respectively. 
We have trained $\mathcal{F}_P$ (a multi-label ResNet50 classifier) on the raw images of the train set and evaluated it on the anonymized version of the test set images. 
We have assessed the PD only on the human subjects that are successfully segmented by the YOLO to reduce the impact of mis-detections, such as heavily occluded individuals or small faces.

Table~\ref{tbl:pd_vispr_clf_compare} presents the quantitative performance of the $\mathcal{F}_P$ that the baseline ANs provide $5.02\%$ to $33.38\%$ reduction in mAP and $12.60\%$ to $86.43\%$ in mR, whereas our adaptive ANs achieve $27.35\%$ to $40.40\%$ in mAP and $77.71\%$ to $88.53\%$ in mR. The relatively higher mR than the mAP drop indicates that the AN reduces confidence of the PD and results in lower recall at the threshold of $0.5$. The $\mathcal{P}^\text{0}_{\mathbf{d}\in\{2,4\}}$ ANs, with fixed factors, are inadequate. 
Our adaptive ANs have enhanced the AN robustness compared to their corresponding non-adaptive baselines. 
We have leveraged the $\mathcal{G}$ by $7.36\%\text{--}13.60\%$ in mAP, and $54.19\%\text{--}64.04\%$ in mR, and the $\mathcal{P}$ by $7.34\%\text{--}35.98\%$ in mAP, and $42.06\%\text{--}83.86\%$ in mR. 

\begin{figure}[htbp]
\centering
\begin{subfigure}[]{1\columnwidth}
\centering
\includegraphics[width=1\linewidth]{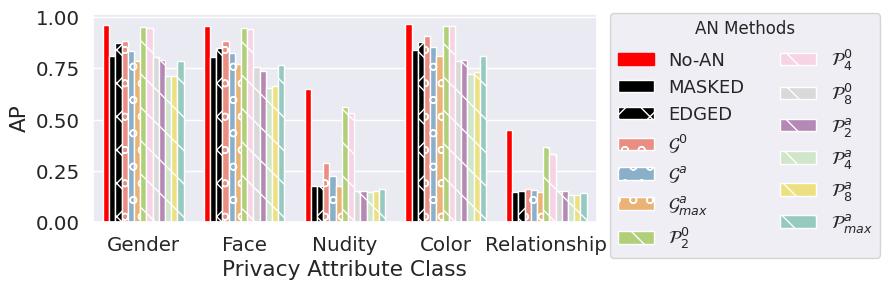}
\end{subfigure}
\caption{{PD per privacy attribute on the VISPR dataset~\cite{orekondy2017towards}.}}
\label{fig:pd_vispr_clf_compare_per_class}
\vspace*{-0.5\baselineskip}
\end{figure}

Fig.~\ref{fig:pd_vispr_clf_compare_per_class} illustrates class-wise AP and R, portraying the impact of different ANs on the privacy attribute classes. 
The lighter non-adaptive ANs with $\mathcal{P}^\text{0}_{\mathbf{d}\in\{2,4\}}$ have delivered a weak privacy shield across all attributes, including nudity and relationship, where the other methods have consistently accomplished improved protection.  
The bounded ANs with $\mathcal{P}^\text{a}$ offer superior protection compared to $\mathcal{P}^\text{0}$, MASKED, and blurring ANs. 
Our adaptive ANs have significantly improved performance, achieving better results using blurring and pixelization (even at lower $\mathbf{d}\in\{2,4\}$), compared to the baselines.
The $\mathcal{G}^\text{a}_\text{max}$, i.e., $\mathbf{k^a_i} = \mathbf{z_i}$, enhances the AN against gender and face detection than the bounded AN $\mathcal{G}^\text{a}$, which regulates the $\mathbf{k^a_i}$ through $\alpha_l \in \{0.5, 1.0\}$. 
However, the $\mathcal{P}^\text{a}_\text{max}$, i.e., $\mathbf{d^a_i} = \mathbf{z_i}$, which results in a colored silhouette, provides a weaker AN than the bounded AN with $\mathcal{P}^\text{a}_{{d}\in\{4,8\}}$. The AN declines for $\mathcal{P}^\text{a}_\text{max}$ since the AP raises by $\Delta \text{AP}_\text{gender}=29\%$, $\Delta \text{AP}_\text{face}=33\%$ and $\Delta \text{AP}_\text{color}=36\%$ as compared to the $\mathcal{P}^\text{a}_{{d}\in\{4,8\}}$.
The results demonstrate that AN deteriorates when the mask color is inferred from the original image pixels. 

The full-body MASKED AN falls short of achieving strong protection against all privacy attributes, particularly for gender, face, and skin color, despite common perceptions and expectations. 
The EDGED, adding edge lines to the MASKED, further deteriorates the protection. 
The leakage on the face and gender can pertain to the $\mathcal{F_P}$ model, exploiting body shapes (contours) for attribute detection.
However, we hypothesize that the ineffectiveness against the skin color is influenced by the prominent association of color class label with gender and facial labels (as shown in Fig.~\ref{fig:pd_vispr_multi__train_labels}).
Our PD assessment on the VISPR dataset aligns with the results of Refs.~\cite{dave2022spact, fioresi2023ted}. But, the strong label co-occurrences have affected the classification evaluation among the face, gender, and skin color classes, and eventually, limit the overall PD competency of the AN. In our study, we further examine the AN using ReID on the Market1501 dataset in Section~\ref{sec:results_reid}.

\begin{table}[htbp] \small 
\centering
\caption{PD Evaluation on the VISPR Dataset~\cite{orekondy2017towards} using \textsc{ResNet50}. For No-AN+: $\forall c_i$, and Others: $\forall c_{i\neq1}$ to Analyze the Effect of AN only on the Privacy Attributes.}
\resizebox{1\columnwidth}{!}{
\begin{tabular}{lcccccc}
\hline
AN Method & mAP$\downarrow$  &    $\Delta_\text{N}\downarrow$  &    $\Delta_\text{0}\downarrow$     &   mR$\downarrow$   &  $\Delta_\text{N}\downarrow$  &    $\Delta_\text{0}\downarrow$    \\ \hline
No-AN+                             & 0.824            & --                               & --                               & 0.665           & --                                  & --                                  \\
No-AN                              & 0.797            & --                               & --                               & 0.619           & --                                  & --                                  \\
MASKED                               & 0.556            & -30.24\%                         & --                               & 0.084           & -86.43\%                            & --                                  \\
EDGED                                   & 0.587            & -26.35\%                         & --                               & 0.191           & -69.14\%                            & --                                  \\
$\mathcal{G}^\text{0}$                      & 0.625            & -21.58\%                         & --                               & 0.203           & -67.21\%                            & --                                  \\
$\mathcal{G}^\text{a} (\alpha_l=0.5)$         & 0.579            & -27.35\%                         & -7.36\%                          & 0.093           & -84.98\%                            & -54.19\%                            \\
$\mathcal{G}^\text{a} (\alpha_l=1.0)$         & 0.575            & -27.85\%                         & -8.00\%                          & 0.090           & -85.46\%                            & -55.67\%                            \\
$\mathcal{G}^\text{a} _\text{max}$   & 0.540            & -32.25\%                         & -13.60\%                         & 0.073           & -88.21\%                            & -64.04\%                            \\
$\mathcal{P}^\text{0}_2$                    & 0.757            & -5.02\%                          & --                               & 0.541           & -12.60\%                            & --                                  \\
$\mathcal{P}^\text{0}_4$                    & 0.742            & -6.90\%                          & --                               & 0.502           & -18.90\%                            & --                                  \\
$\mathcal{P}^\text{0}_8$                    & 0.531            & -33.38\%                         & --                               & 0.126           & -79.64\%                            & --                                  \\
$\mathcal{P}^\text{a}_2(\alpha_l=0.5)$        & 0.524            & -34.25\%                         & -30.78\%                         & 0.138           & -77.71\%                            & -74.49\%                            \\
$\mathcal{P}^\text{a}_4(\alpha_l=0.5)$        & \textbf{0.475}   & \textbf{-40.40\%}                & \textbf{-35.98\%}                & 0.082           & -86.75\%                            & -83.67\%                   \\
$\mathcal{P}^\text{a}_8(\alpha_l=0.5)$        & 0.479            & -39.90\%                         & -9.79\%                          & \textbf{0.071}  & \textbf{-88.53\%}                   & -43.65\%                            \\
$\mathcal{P}^\text{a}_2(\alpha_l=1.0)$        & 0.524            & -34.25\%                         & -30.78\%                         & 0.138           & -77.71\%                            & -74.49\%                            \\
$\mathcal{P}^\text{a}_4(\alpha_l=1.0)$        & \textbf{0.475}   & \textbf{-40.40\%}                & \textbf{-35.98\%}                & 0.081           & -86.91\%                            & \textbf{-83.86\%}                   \\
$\mathcal{P}^\text{a}_8(\alpha_l=1.0)$        & 0.492            & -38.27\%                         & -7.34\%                          & 0.073           & -88.21\%                            & -42.06\%                            \\
$\mathcal{P}^\text{a} _\text{max}$   & 0.533            & -33.12\%                         & --                               & 0.082           & -86.75\%                            & --                                  \\
*\textsc{SPAct}~\cite{dave2022spact} & 0.527 & -15.39\%  & -- & -- &  -- & --  \\ 
*\textsc{TeD-SPAD}~\cite{fioresi2023ted} &  0.422 & -32.25\%  &  --& -- &  -- & --  \\ \hline
\end{tabular}
}
\noindent\footnotesize{\par  
The * denotes the reported scores of DL-based full-image AN methods in Ref.~\cite{fioresi2023ted}, where the $\text{mAP} = 0.623$ on the No-AN.
\par}
\label{tbl:pd_vispr_clf_compare}
\vspace*{-0.5\baselineskip}
\end{table}

Table~\ref{tbl:vispr_pd_model_compare_loss} presents an ablation study on the $\mathcal{F}_P$ model to demonstrate the relative gain of our proposed fine-tuning and weighted loss mechanisms. Employing a pre-trained \textsc{ResNet50} model for initialization reduces the test MSE loss by $44.1\%$ and improves the mAP by $25.5\%$ as compared to the baseline approaches employed in Ref.~\cite{dave2022spact, fioresi2023ted}. The weighted loss, given in Eq.~\eqref{eq:weighted_loss_fn}, further increases the mAP by $3.3\%$ and leverages the attributes labels with relatively lower contributions, i.e., nudity and relationship classes; the AP has been increased by $1.26\%$ and $9.79\%$, respectively.

\begin{table}[htbp]
\centering
\caption{Ablation Study of $\mathcal{F}_P$ on the Non-Anonymized Test Data. }
\resizebox{1\columnwidth}{!}{
\begin{tabular}{lcccc}
\hline
{Model ($\mathcal{F}_P$)}  & {Initializing} & {Training} & {Test MSE $\mathcal{L}$$\downarrow$} & {mAP$\uparrow$} \\ \hline
\textsc{ResNet50} (\cite{dave2022spact, fioresi2023ted})                              & Random & $\mathcal{L}$                    &       0.381             &     0.652          \\ 
\textsc{ResNet50} (ours)             & pre-trained~\cite{he2016deep} &  $\mathcal{L}$                    & 0.213 ($44.1\%\downarrow$)           & 0.818 ($25.5\%\uparrow$)      \\ 
\textsc{ResNet50} (ours)            &  pre-trained~\cite{he2016deep} &  $\mathcal{L}_w$                 & \textbf{0.206} (\textbf{45.9\%}$\downarrow$)          & \textbf{0.824} (\textbf{26.4\%}$\uparrow$)      \\ \hline
\end{tabular}
}
\par \footnotesize 
The loss $\mathcal{L}$ and $\mathcal{L}_w$ denote without and with class weighting, respectively. 
\par
\label{tbl:vispr_pd_model_compare_loss}
\vspace*{-\baselineskip}
\end{table}

\subsubsection{Person Re-Identification}
\label{sec:results_reid}

We evaluate person recognition rate from full-body images on the Market1501 dataset~\cite{zheng2015scalable}. 
The dataset contains a cross-camera ReID benchmark data from six cameras, including five high-resolution cameras and one low-resolution camera, capturing 1,501 identities with overlapping fields of view. It comprises a total of 32,668 annotated bounding boxes.    
We adopt the \textsc{OSNet}~\cite{zhou2021learning}, a state-of-the-art ReID model, for our experiment. The \textsc{OSNet} computes Euclidean distance on the 512 embedded features of the query and gallery images to evaluate matching. We have applied reversible padding to resize the images from $[64, 128]$ to YOLO's input dimension of $[320, 240]$, thereby enhancing object segmentation accuracy. We have found that the padding considerably reduces the person detection missing rate from $8.5\%$ to $0.62\%$ on the query image set as compared to direct resizing with and without dimension ratio preservation. 

Table~\ref{tbl:reid_merket1501_clf_compare_all} presents the ReID performance when the AN is applied to both the query and gallery images. The reference mAP and CMC-R1 of the \textsc{OSNet} on the non-anonymized images are $0.83$ and $0.94$, respectively. 
The MASKED AN drops the mAP and the CMC-R1 by $98.2\%$ and $94.9\%$, respectively, accomplishing a strong protection. 
The other baseline methods yield around $60\%$ reduction in the mAP except for the $\mathcal{P}^\text{0}_4$ and $\mathcal{P}^\text{0}_2$, achieving only $21.5\%$ and $3.4\%$, respectively. 
The results demonstrate that $\mathcal{P}$ does not adequately prevent the ReID unless a higher downsizing factor $\mathbf{d}$ is employed. 
Our adaptive $\mathcal{P}^\text{a}_{\{2,4\}}$ have relatively improved the $\mathcal{P}^\text{0}_{\{2,4\}}$, i.e., the mAP by $94.4\%$ and $91.1\%$, and the CMC-R1 by $87.2\%$ and $85.8\%$, respectively, with $\alpha_l=0.5$. We have also reduced the performance gap among the $\mathcal{P}$ while enhancing the mAP of the $\mathcal{G}$ by $48.4\%\text{--}66.0\%$. 
We provide further ReID discussion in Appendix~\ref{sec:results_reid_extra} when the recognition attack employs non-anonymized data on either the query or the search gallery.

\begin{table}[!t]
\centering
\caption{ReID (Anonymized Query vs.  Anonymized Gallery) on the Market1501 Dataset~\cite{zheng2015scalable} using \textsc{OSNet}~\cite{zhou2021learning}.}
\resizebox{1\columnwidth}{!}{
\begin{tabular}{lcccccc}
\hline
AN Method                                          & mAP$\downarrow$ & $\Delta_\text{N}\downarrow$ & $\Delta_\text{0}\downarrow$ & CMC-R1$\downarrow$ & $\Delta_\text{N}\downarrow$ & $\Delta_\text{0}\downarrow$ \\ \hline
No-AN                                      & 0.826           & --                                  & --                              & 0.942              & --                                     & --                                     \\
MASKED                                       & \textbf{0.015}  & \textbf{-98.2\%}                    & --                              & \textbf{0.048}     & \textbf{-94.9\%}                       & --                                     \\
EDGED                                           & 0.016           & -98.1\%                             & --                              & \textbf{0.048}     & \textbf{-94.9\%}                       & --                                     \\
$\mathcal{G}^\text{0}$                          & 0.335           & -59.4\%                             & --                              & 0.596              & -36.7\%                                & --                                     \\
$\mathcal{G}^\text{a} (\alpha_l=0.5)$           & 0.173           & -79.1\%                             & -48.4\%                         & 0.370              & -60.7\%                                & -37.9\%                                \\
$\mathcal{G}^\text{a} (\alpha_l=1.0)$           & 0.154           & -81.4\%                             & -54.0\%                         & 0.340              & -63.9\%                                & -43.0\%                                \\
$\mathcal{G}^\text{a} _\text{max}$ & 0.114           & -86.2\%                             & -66.0\%                         & 0.253              & -73.1\%                                & -57.6\%                                \\
$\mathcal{P}^\text{0}_2$                        & 0.798           & -3.4\%                              & --                              & 0.928              & -1.5\%                                 & --                                     \\
$\mathcal{P}^\text{0}_4$                        & 0.648           & -21.5\%                             & --                              & 0.839              & -10.9\%                                & --                                     \\
$\mathcal{P}^\text{0}_8$                        & 0.307           & -62.8\%                             & --                              & 0.538              & -42.9\%                                & --                                     \\
$\mathcal{P}^\text{a}_2(\alpha_l=0.5)$          & 0.045           & -94.6\%                             & -94.4\%                         & 0.119              & -87.4\%                                & -87.2\%                                \\
$\mathcal{P}^\text{a}_4(\alpha_l=0.5)$          & 0.056           & -93.2\%                             & -91.4\%                         & 0.119              & -87.4\%                                & -85.8\%                                \\
$\mathcal{P}^\text{a}_8(\alpha_l=0.5)$          & 0.058           & -93.0\%                             & -81.1\%                         & 0.126              & -86.6\%                                & -76.6\%                                \\
$\mathcal{P}^\text{a}_2(\alpha_l=1.0)$          & 0.040           & -95.2\%                             & \textbf{-95.0\%}                & 0.105              & -88.9\%                                & -88.7\%                                \\
$\mathcal{P}^\text{a}_4(\alpha_l=1.0)$          & 0.035           & -95.8\%                             & -94.6\%                         & 0.093              & -90.1\%                                & \textbf{-88.9\%}                       \\
$\mathcal{P}^\text{a}_8(\alpha_l=1.0)$          & 0.035           & -95.8\%                             & -88.6\%                         & 0.080              & -91.5\%                                & -85.1\%                                \\
$\mathcal{P}^\text{a} _\text{max}$ & 0.035           & -95.8\%                             & --                              & 0.080              & -91.5\%                                & --        \\ \hline                           
\end{tabular}
}
\label{tbl:reid_merket1501_clf_compare_all}
\vspace*{-1\baselineskip}
\end{table}

\begin{figure}[!htbp]
\centering
\begin{subfigure}[]{0.75\columnwidth}
\centering
\begin{tabular}{c@{}c@{ }c@{}c@{ }c@{}c}
\includegraphics[width=0.16\linewidth]{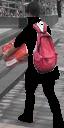} &
\includegraphics[width=0.16\linewidth]{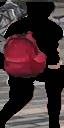} &
\includegraphics[width=0.16\linewidth]{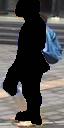} &
\includegraphics[width=0.16\linewidth]{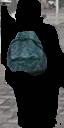} &
\includegraphics[width=0.16\linewidth]{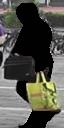} &
\includegraphics[width=0.16\linewidth]{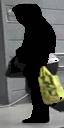}
\end{tabular}
\caption{}
\end{subfigure}
\begin{subfigure}[]{0.75\columnwidth}
\centering
\begin{tabular}{c@{}c@{ }c@{}c@{ }c@{}c}
\includegraphics[width=0.16\linewidth]{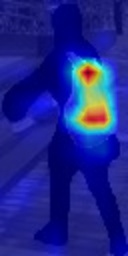} &
\includegraphics[width=0.16\linewidth]{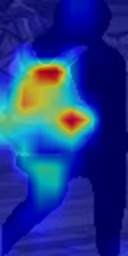} &
\includegraphics[width=0.16\linewidth]{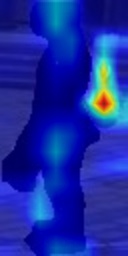} &
\includegraphics[width=0.16\linewidth]{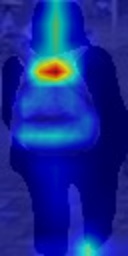} &
\includegraphics[width=0.16\linewidth]{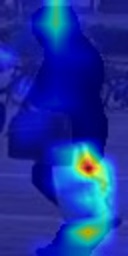} &
\includegraphics[width=0.16\linewidth]{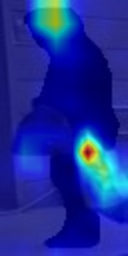}
\end{tabular}
\caption{}
\end{subfigure}
\caption{Sample images with a person holding objects, ReID leakage after AN: a) paired images are sampled from the query and gallery, and b) class activation maps (the red indicates high score regions), showing the ReID's focus on the objects.
}
\label{fig:reid_merket1501_anony_persnalitems_samples}
\vspace*{-1.5\baselineskip}
\end{figure}

Despite the YOLO's $99.3\%$ person detection accuracy on the Market1501 dataset, the MASKED AN achieves mAP of $0.015$ ($98.2\%\downarrow$). Several factors can contribute to the small leakage, such as undetected human subjects, incomplete segmentation, and non-anonymized items the person is holding, like bags. While the former two reasons mainly correspond to poor image qualities, occlusions, and model accuracy, we found identity leakage through personal item characteristics very interesting (see Fig.~\ref{fig:reid_merket1501_anony_persnalitems_samples}). When personal objects, such as handbags, backpacks, umbrellas, suitcases, cellphones, and laptops, are anonymized together with the person's body, the AN improves as the mAP falls further to $0.006$ ($99.27\%\downarrow$).
The mAP also decreases (improves) for EDGED:~$0.016 \shortrightarrow 0.013$, 
$\mathcal{G}^\text{0}$:~$0.335 \shortrightarrow 0.318$, 
$\mathcal{G}^\text{a}$:~$0.173 \shortrightarrow 0.160$, 
$\mathcal{G}^\text{a} _\text{max}$:~$0.114 \shortrightarrow 0.103$,
$\mathcal{P}^\text{0}_2$:~$0.798 \shortrightarrow 0.798$, 
$\mathcal{P}^\text{0}_4$:~$0.648 \shortrightarrow 0.643$, 
$\mathcal{P}^\text{0}_8$:~$0.307 \shortrightarrow 0.294$, 
$\mathcal{P}^\text{a}_2$:~$0.045 \shortrightarrow 0.042$, 
$\mathcal{P}^\text{a}_4$:~$0.056 \shortrightarrow 0.053$, 
$\mathcal{P}^\text{a}_8$:~$0.058 \shortrightarrow 0.055$, and 
$\mathcal{P}^\text{a} _\text{max}$:~$0.035 \shortrightarrow 0.032$.

\subsection{Evaluation on Video Anomaly Detection}

We evaluate the utility quality of the AN for the VAD using large-scale public datasets, i.e., the UCF-Crime~\cite{sultani2018real} and XD-Violence~\cite{Wu2020not}. The UCF-Crime dataset comprises 128 hours of untrimmed CCTV videos from 1,900 sources~\cite{sultani2018real}. The videos capture 13 crime anomaly types featuring real-world incidents. The test set comprises 290 videos with varying frame lengths, totaling over 1 million frames. The XD-Violence dataset, the largest multi-scene weakly labeled VAD dataset to date, comprises a total duration of 217 hours from 4754 videos~\cite{Wu2020not}. Its test set includes 800 videos, totaling over 2.3 million frames, and covers 6 anomaly categories related to violence. The videos were collected from a variety of sources, including different types of cameras, movies, and games, which creates a mix of scenes that increases the VAD complexity. 
We adopt the \textsc{PEL4VAD}~\cite{pu2024learning} and \textsc{MGFN}~\cite{chen2023mgfn} VAD models for the evaluation. The models were trained on \textsc{I3D-RGB} features extracted from the raw videos of the UCF-Crime and XD-Violence datasets. We provide details of the video encoding mechanism in Appendix~\ref{sec:vad_i3d_encoding}.

\begin{figure}[t]
\centering
\includegraphics[width=1\columnwidth]{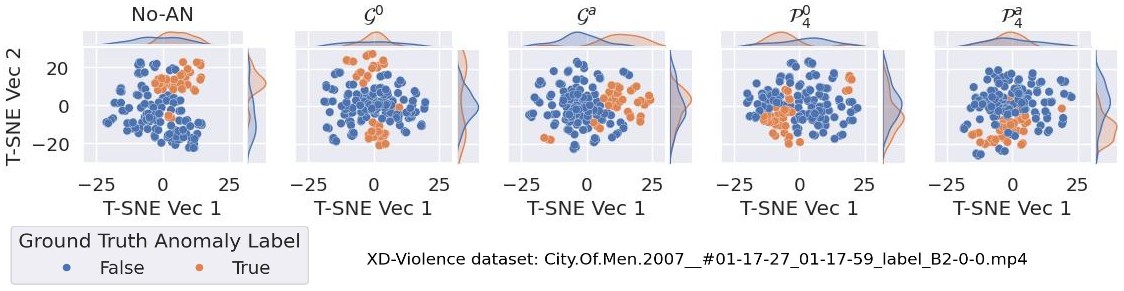} 
\caption{Illustration anomaly localization robustness using 2D T-SNE embedding of the video I3D features, demonstrating the $\mathcal{G}^\text{a}$ improves the feature distribution clustering compared to $\mathcal{G}^\text{0}$. The $\mathcal{P}^\text{a}_4$ adds a slight perturbation that lowers the VAD.
}
\label{fig:pel_ad_xd_auc_i3d_tnse_compare_fCity_Of_Men_2007__011727_011759_label_B200}
\vspace*{-1\baselineskip}
\end{figure}

Table~\ref{tbl:ad_pel4vad_mgfn_ucf_clf_comapre} provides the VAD AUC performance on the test set of the UCF-Crime dataset. {To optimize the paper length, we limit the discussion henceforth to the bounded $\Theta^\text{a}$ using $\alpha_r=1.0$ and $\alpha_l=0.5$.} 
The \textsc{PEL4VAD} outperforms the \textsc{MGFN} in all scenarios, with an AUC of $0.86$ on the raw videos compared to $0.83$ for the \textsc{MGFN}. 
The AN degrades the VAD competency by $0.00\%\text{--}2.90\%$ for the \textsc{PEL4VAD} (except for the $\mathcal{P}^\text{0}_2$ with a slight positive gain by $+0.23\%$) and by $0.00\%\text{--}4.83\%$ for the \textsc{MGFN} (except for the $\mathcal{P}^\text{0}_2$ with $+0.60\%$ and MASKED with $+0.36\%$). 
Our \textsc{LA3D} approach improves the VAD with $\mathcal{G}^\text{a}$ for both models, the \textsc{PEL4VAD} by $0.12\%\text{--}0.23\%$, and \textsc{MGFN} by $1.10\%$, as compared to their corresponding non-adaptive methods. It provides slightly lower for $\mathcal{P}^\text{a}$, the \textsc{PEL4VAD} by $0.23\%\text{--}0.83\%$ and \textsc{MGFN} by $0.60\%\text{--}2.42\%$ with $+1.00\%$ positive gain for the $\mathcal{P}^\text{a}_8$. 
Table~\ref{tbl:ad_pel4vad_mgfn_xd_clf_comapre} presents the VAD performance on the XD-Violence dataset, where the AUC drop for the \textsc{PEL4VAD} is $0.00\%\text{--}3.29\%$ and $0.00\%\text{--}4.37\%$ for \textsc{MGFN}. The EDGED AN is an outlier for the \textsc{PEL4VAD} with a drop of $5.53\%$. The XD-Violence contains higher-quality (resolution) videos than the UCF-Crime dataset, and heavy ANs, such as MASKED and EDGED, have a greater deterioration effect on the VAD utility. 
The $\mathcal{G}^\text{a}$ results in positive VAD gain, but $\mathcal{P}^\text{a}_\mathbf{d}$ can have a slight negative impact depending on $\mathbf{d}$. We hypothesize that a higher smoothing effect of the $\mathcal{G}^\text{a}$ can denoise irrelevant details in the data and improve the VAD models, whereas bounded $\mathcal{P}^\text{a}$ increases perturbation. Fig.\ref{fig:pel_ad_xd_auc_i3d_tnse_compare_fCity_Of_Men_2007__011727_011759_label_B200} shows a 2D embedding of the I3D features of a sample video to demonstrate the robustness of the ANs in anomaly localization.
Our \textsc{LA3D} approaches have significantly outperformed benchmark DL models, such as \textsc{SPAct}~\cite{dave2022spact} and \textsc{TeD-SPAD}~\cite{fioresi2023ted}, on both datasets.

\begin{table}[htbp]
\centering
\caption{VAD on UCF Dataset~\cite{sultani2018real} using \textsc{PEL4VAD}~\cite{pu2024learning} and \textsc{MGFN}~\cite{chen2023mgfn}.}
\resizebox{1\columnwidth}{!}{
\begin{tabular}{lcccccc}
\hline
AN Method                                & AUC$_\text{PEL4VAD}\uparrow$ & $\Delta_\text{N}\uparrow$ & $\Delta_\text{0}\uparrow$ & AUC$_\text{MGFN}\uparrow$ & $\Delta_\text{N}\uparrow$ & $\Delta_\text{0}\uparrow$ \\ \hline
No-AN                                                                                                         & 0.863                        & --                        & --                        & 0.830                           & --                                & --                        \\
MASKED                                                                                                        & 0.861                        & -0.23\%                   & --                        & 0.833                           & 0.36\%                              & --                        \\
EDGED                                                                                                         & 0.849                        & -1.62\%                   & --                        & 0.826                           & -0.48\%                             & --                        \\
$\mathcal{G}^\text{0}$                                                                                        & 0.857                        & -0.70\%                   & --                        & 0.816                           & -1.69\%                             & --                        \\
$\mathcal{G}^\text{a}(\alpha_l=0.5)$                                                                          & 0.859                        & -0.46\%                   & \textbf{0.23\%}           & 0.825                           & -0.60\%                             & \textbf{1.10\%}           \\
$\mathcal{G}^\text{a} _\text{max}$                                                                            & 0.858                        & -0.58\%                   & 0.12\%                    & 0.825                           & -0.60\%                             & \textbf{1.10\%}           \\
$\mathcal{P}^\text{0}_2$                                                                                      & \textbf{0.865}               & \textbf{0.23\%}           & --                        & \textbf{0.835}                  & \textbf{0.60\%}                     & --                        \\
$\mathcal{P}^\text{0}_4$                                                                                      & 0.853                        & -1.16\%                   & --                        & 0.825                           & -0.60\%                             & --                        \\
$\mathcal{P}^\text{0}_8$                                                                                      & 0.845                        & -2.09\%                   & --                        & 0.802                           & -3.37\%                             & --                        \\
$\mathcal{P}^\text{a}_2(\alpha_l=0.5)$                                                                        & 0.863                        & 0.00\%                    & -0.23\%                   & 0.830                           & 0.00\%                              & -0.60\%                   \\
$\mathcal{P}^\text{a}_4(\alpha_l=0.5)$                                                                        & 0.851                        & -1.39\%                   & -0.23\%                   & 0.805                           & -3.01\%                             & -2.42\%                   \\
$\mathcal{P}^\text{a}_8(\alpha_l=0.5)$                                                                        & 0.838                        & -2.90\%                   & -0.83\%                   & 0.810                           & -2.41\%                             & 1.00\%                    \\
$\mathcal{P}^\text{a} _\text{max}$                                                                            & 0.860                        & -0.35\%                   & --                        & 0.827                           & -0.36\%                             & --                        \\
*\textsc{SPAct}~\cite{dave2022spact}     & --                           & --                      & --                      & 0.739        & -4.83\%    & --                      \\
*\textsc{TeD-SPAD}~\cite{fioresi2023ted} & --                           & --                      & --                      & 0.748        & -3.69\%    & --    \\ \hline
\end{tabular}
}
\label{tbl:ad_pel4vad_mgfn_ucf_clf_comapre}
\par \footnotesize  
The * denotes the reported scores of DL-based full-image AN methods in Ref.~\cite{fioresi2023ted}, where the $\text{AUC} = 0.777$ on the No-AN. 
\par
\vspace*{-0.5\baselineskip}
\end{table}

\begin{table*}[htbp]
\centering
\caption{VAD on XD Violence Dataset~\cite{Wu2020not} using \textsc{PEL4VAD}~\cite{pu2024learning} and \textsc{MGFN}~\cite{chen2023mgfn}.}
\resizebox{0.9\textwidth}{!}{
\begin{tabular}{lcccccccccccc}
\hline
AN Method                                & AUC$_\text{PEL4VAD}\uparrow$ & $\Delta_\text{N}\uparrow$ & $\Delta_\text{0}\uparrow$ & AP$_\text{PEL4VAD}\uparrow$ & $\Delta_\text{N}\uparrow$ & $\Delta_\text{0}\uparrow$ & AUC$_\text{MGFN}\uparrow$ & $\Delta_\text{N}\uparrow$ & $\Delta_\text{0}\uparrow$ & AP$_\text{MGFN}\uparrow$ & $\Delta_\text{N}\uparrow$ & $\Delta_\text{0}\uparrow$ \\ \hline
No-AN                                                                                                         & \textbf{0.941} & --    & --    & \textbf{0.832} & --     & --    & \textbf{0.915} & --    & --    & \textbf{0.763} & --     & --     \\
MASKED                                                                                                        & 0.910 & -3.29\% & --    & 0.761 & -8.53\%  & --    & 0.877 & -4.15\% & --    & 0.672 & -11.93\% & --     \\
EDGED                                                                                                         & 0.889 & -5.53\% & --    & 0.705 & -15.26\% & --    & 0.876 & -4.26\% & --    & 0.678 & -11.14\% & --     \\
$\mathcal{G}^\text{0}$                                                                                        & 0.932 & -0.96\% & --    & 0.817 & -1.80\%  & --    & 0.901 & -1.53\% & --    & 0.750 & -1.70\%  & --     \\
$\mathcal{G}^\text{a}(\alpha_l=0.5)$                                                                          & 0.931 & -1.06\% & -0.11\% & 0.819 & -1.56\%  & 0.24\%  & 0.906 & -0.98\% & 0.55\%  & 0.756 & -0.92\%  & \textbf{0.80\%}   \\
$\mathcal{G}^\text{a} _\text{max}$                                                                            & 0.933 & -0.85\% & 0.11\%    & 0.822 & -1.20\%  & 0.61\%    & 0.910 & -0.55\% & \textbf{1.00\%}   & 0.753 & -1.31\%  & 0.40\%    \\
$\mathcal{P}^\text{0}_2$                                                                                      & \textbf{0.941} & \textbf{0.00\%}  & --    & \textbf{0.832} & \textbf{0.00\%}   & --    & \textbf{0.915} & \textbf{0.00\%}  & --    & \textbf{0.763} & \textbf{0.00\%}   & --     \\
$\mathcal{P}^\text{0}_4$                                                                                      & 0.927 & -1.49\% & --    & 0.807 & -3.00\%  & --    & 0.894 & -2.30\% & --    & 0.721 & -5.50\%  & --     \\
$\mathcal{P}^\text{0}_8$                                                                                      & 0.911 & -3.19\% & --    & 0.768 & -7.69\%  & --    & 0.875 & -4.37\% & --    & 0.686 & -10.09\% & --     \\
$\mathcal{P}^\text{a}_2(\alpha_l=0.5)$                                                                        & 0.925 & -1.70\% & -1.70\% & 0.802 & -3.61\%  & -3.61\% & 0.896 & -2.08\% & -2.08\% & 0.729 & -4.46\%  & -4.46\%  \\
$\mathcal{P}^\text{a}_4(\alpha_l=0.5)$                                                                        & 0.919 & -2.34\% & -0.86\% & 0.790 & -5.05\%  & -2.11\% & 0.893 & -2.40\% & -0.11\% & 0.716 & -6.16\%  & -0.69\%  \\
$\mathcal{P}^\text{a}_8(\alpha_l=0.5)$                                                                        & 0.917 & -2.55\% & \textbf{0.66\%}  & 0.775 & -6.85\%  & \textbf{0.91\%}  & 0.880 & -3.83\% & 0.57\%  & 0.680 & -10.88\% & -0.87\%  \\
$\mathcal{P}^\text{a} _\text{max}$                                                                            & 0.931 & -1.06\% & --    & 0.816 & -1.92\%  & --    & 0.909 & -0.66\% & --    & 0.751 & -1.57\%  & --     \\
{*\textsc{SPAct}~\cite{dave2022spact}}     & --    & --    & --    & --    & --     & --    & --    & --    & --    & 0.534 & --     & -27.62\% \\
{*\textsc{TeD-SPAD}~\cite{fioresi2023ted}} & --    & --    & --    & --    & --     & --    & --    & --    & --    & 0.603 & --     & -18.18\%  \\ \hline
\end{tabular}
}
\label{tbl:ad_pel4vad_mgfn_xd_clf_comapre}
\footnotesize{\par 
The * denotes the reported scores of DL-based full-image AN methods in Ref.~\cite{fioresi2023ted}, where the $\text{AP} = 0.737$ on the No-AN. \par}
\end{table*}

\begin{figure*}[htbp]
\centering
\begin{subfigure}[]{0.48\textwidth}
\includegraphics[width=1\columnwidth]{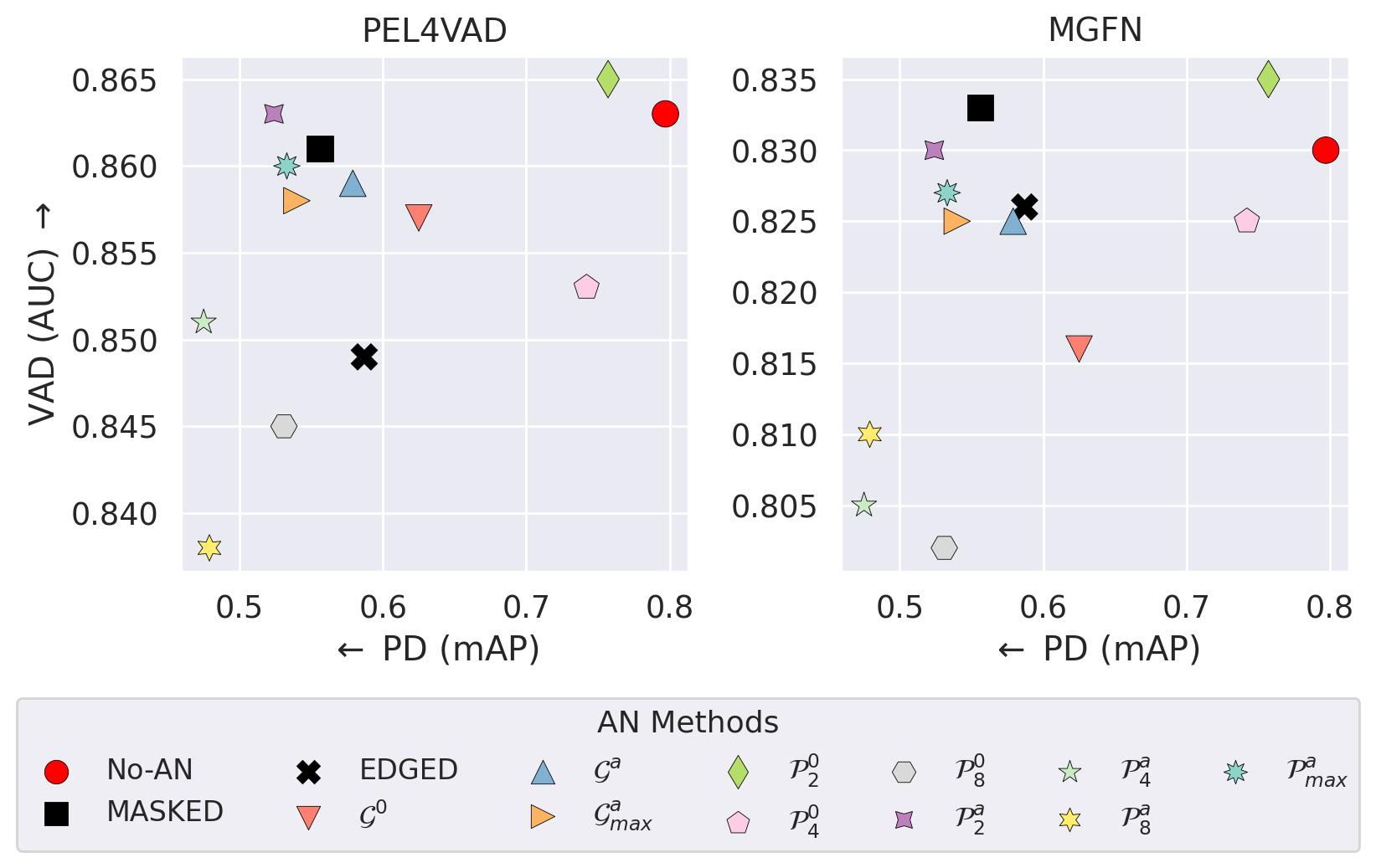} 
\caption{}
 \end{subfigure}
 \begin{subfigure}[]{0.48\textwidth}
\includegraphics[width=1\columnwidth]{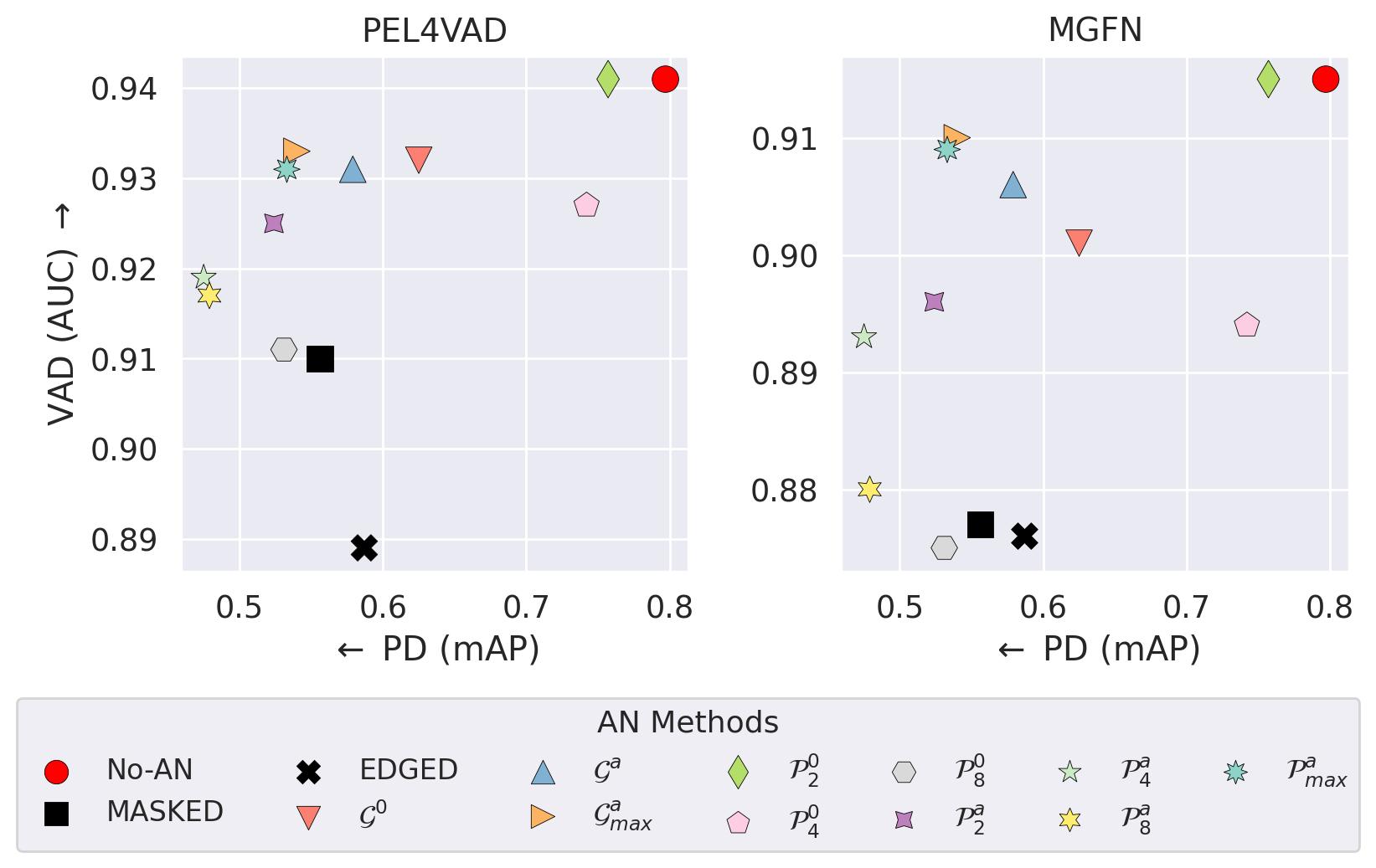} 
\caption{}
 \end{subfigure}
\caption{Performance of AN in privacy protection (PD on the VISPR dataset~\cite{orekondy2017towards} from Table~\ref{tbl:pd_vispr_clf_compare}) vs. VAD utility preservation: a) UCF-Crime dataset~\cite{sultani2018real} from Table~\ref{tbl:ad_pel4vad_mgfn_ucf_clf_comapre}, and b) XD-Violence dataset~\cite{sultani2018real} from Table~\ref{tbl:ad_pel4vad_mgfn_xd_clf_comapre}.
}
\label{fig:ad_ucf_auc_pd_vispr_cmap_clf_compare_all}
\vspace*{-0.5\baselineskip}
\end{figure*}

Fig.~\ref{fig:ad_ucf_auc_pd_vispr_cmap_clf_compare_all} illustrates the trade-off of the VAD versus the PD.
We have observed consistent impact patterns of the AN methods across the models and datasets. 
Stronger bounded $\mathcal{P}$ AN tends to decline the VAD, whereas stronger $\mathcal{G}$ improves the VAD. 
For the UCF-Crime dataset, despite the $\mathcal{P}^\text{a}_2$'s good compromise between AN and VAD, the $\mathcal{P}^\text{a}_4$ offers better privacy shield against the human vision (as shown in Figs.~\ref{fig:vispr_anony_compare_all_im_s320_240}~and~\ref{fig:vispr_anony_compare_all_im_scaled_compare_all}). Although MASKED has performed well with both models for the UCF-Crime dataset, it deteriorates the VAD on the XD-Violence. On the other hand, the \textsc{PEL4VAD} model is more sensitive to the EDGED AN than the \textsc{MGFN} across the datasets (see discussion in Appendix~\ref{sec:results_iqa} that the EDGED AN has the highest impact on video quality). This indicates the need for adaptive $\mathcal{P}$ and $\mathcal{G}$ that deliver strong AN without seriously sacrificing the VAD. Our $\mathcal{G}^{a}_\text{max}$ and $\mathcal{P}^{a}_\text{max}$ AN demonstrate a stable trade-off in VAD across models and datasets.
It is important to note that the VAD models were initially trained on No-AN videos but were evaluated using anonymized videos, for which the models were not specifically optimized. Therefore, we recommend training the models on the anonymized data to enhance the VAD further. Nevertheless, we have observed the performance of \textsc{LA3D} encouraging, considering the limited deterioration in the VAD with notable progress in the AN (see Section~\ref{sec:results_pd}~and~\ref{sec:results_reid}).

We present further results in the Appendix: 1) video quality evaluation in Appendix~\ref{sec:results_iqa}, 2) VAD confidence score evaluation and visual illustration in the Appendix~\ref{sec:results_vad_ad_score}, and 3) limitations and future research directions in the Appendix~\ref{sec:results_limitation}.

\subsection{{Computational Cost Analysis}}
\label{sec:results_compcost}

The computational cost of the full-body AN with VAD primarily stems from the target segmentation process, with an average per-frame processing time of $0.011$ seconds, resulting in a cost increase of $12\%$ to $26\%$. 
The bounded adaptive algorithms of the \textsc{LA3D} increase the time by approximately $5\%$ to $6\%$ compared to their non-adaptive settings. 
The AN methods can generally be ranked from fastest to slowest as MASKED, EDGED, $\mathcal{P}^\text{0}$, $\mathcal{G}^\text{0}$, $\mathcal{P}^\text{a}$, and $\mathcal{G}^\text{a}$. 
Incorporating EDGED AN raises the computation by $3\%$. 
{We have also noticed that the computation of $\mathcal{G}$ increases with $\mathbf{k}$ and a slight rise for larger $\mathbf{d}$ of $\mathcal{P}$.}
We conducted our experiment on an Intel(R) Xeon(R) Platinum 8168 CPU @ 2.70GHz with 64GB RAM and Nvidia Tesla V100-SXM3-32GB. 

\begin{figure}[t]
\centering
\includegraphics[width=1\columnwidth]{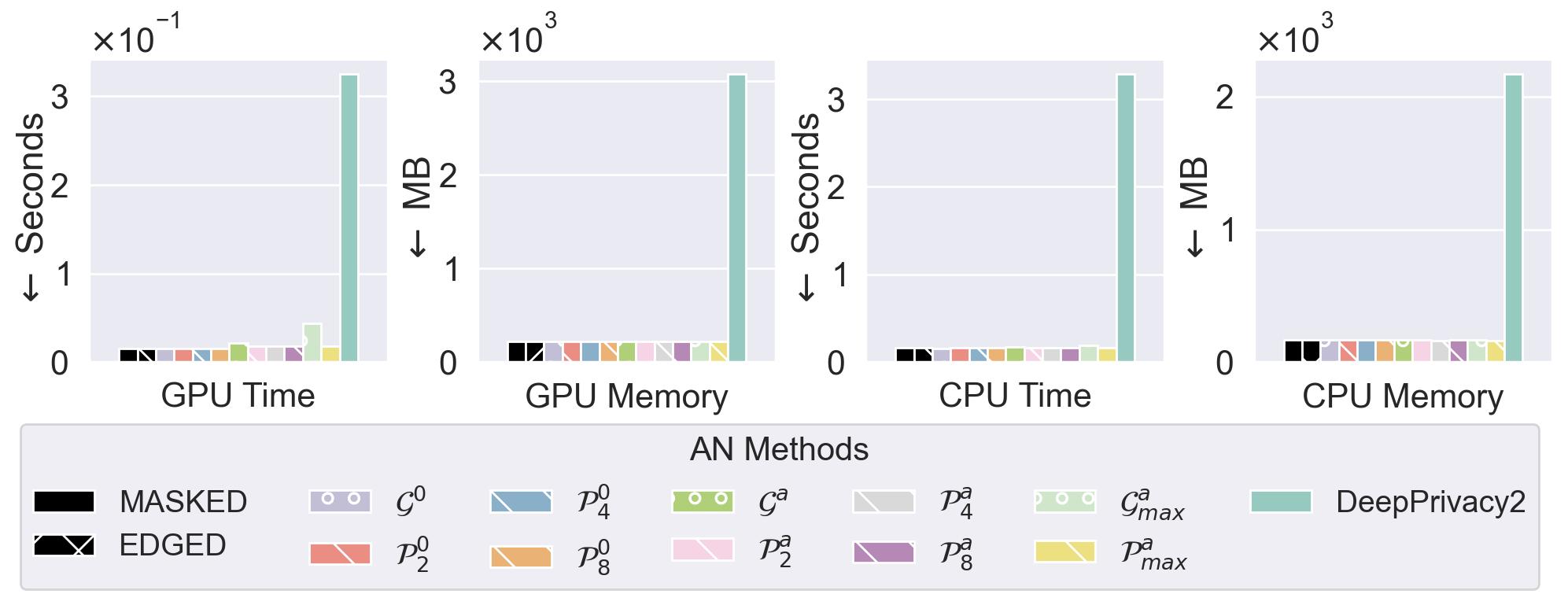} 
\caption{Average per-frame processing time and peak increment memory on the AN on the VISPR dataset using image size of $\mathbf{z} = [320, 240]$. 
The conventional AN employs \textsc{YOLO} models, whereas the \textsc{DeepPrivacy2} employs DL models, such as \textsc{DensePose}, \textsc{MASK-RCNN}, and \textsc{StyleGAN}.
}
\label{fig:cc_vispr_sample_320_240}
\vspace*{-1\baselineskip}
\end{figure}

We have also analyzed the processing time and memory cost (using CPU and GPU modes) of the AN separately on the VISPR dataset and compared them to a recent DL-based (GAN) full-body AN approach, the \textsc{DeepPrivacy2}~\cite{hukkelaas2023deepprivacy2} (recent version of the SG-GAN~\cite{hukkelaas2023realistic}) (see Fig.~\ref{fig:cc_vispr_sample_320_240}). We experimented with two modes: 1) GPU: the deep learning models were executed on the GPU while the others ran on the CPU, and 2) CPU: all algorithms were executed solely on the CPU. The adaptive AN increases the GPU-mode processing time by $3\text{--}6$ milliseconds and $1\text{--}9$ milliseconds for the CPU-mode. But, the $\mathcal{G}^\text{a}_\text{max}$ increases the cost by $30$ milliseconds, as it employs the maximum kernel window size (see Table~\ref {tbl:anony_methods}). The incremental peak memory cost remains roughly 220~MB for the GPU and 160~MB for the CPU, with a negligible difference from the baseline ANs. The DL method has a $16$ times slower speed and a $14$ times higher memory consumption. The \textsc{DeepPrivacy2}, one of the state-of-the-art DL approaches in realistic image generation for AN, considerably sacrifices computation efficiency. The cost analysis demonstrates the feasibility of the proposed lightweight AN approaches for real-time edge CV applications.

\section{{Conclusion}}
\label{sec:conclusion}

In this study, we have conducted experiments on visual anonymization through computationally lightweight conventional algorithms. We have investigated the efficacy trade-off in privacy protection and video anomaly detection using several publicly available benchmark datasets. We have presented elaborate results on the algorithms and empirically demonstrated their potential and limitations.
We have highlighted the limitations of widely utilized conventional and deep learning AN approaches and emphasized the encouraging significance of our simple and efficient adaptive approach in refining the AN against privacy attribute detection, person re-identification, and video anomaly detection.  
Considering the substantial gain on the AN, we have found the performance of the adaptive AN on VAD promising. 
We have demonstrated the promising capability of body-level AN, and pointed out that the performance trade-offs depend on the choice of the AN techniques and the utility models.
The results have established that lightweight AN, with appropriate adaptive enhancements, has potential for privacy-aware real-time and edge computer vision applications. 
The presented baseline performance of the various techniques will provide a good benchmarking reference for future privacy-related studies, further underlining the significance of our study.

\bibliographystyle{IEEEtran}
\footnotesize
\bibliography{bibi.bib}

\appendix
\section*{Supplementary Results}
\label{sec:appendix}

In this appendix, we present further empirical results of the study.
Further supplementary materials on hyperparameters can be found in this  \href{https://github.com/muleina/LA3D/blob/main/results/Supplementary_unpublished.pdf}{link}.

\subsection{{Benchmark Deep Learning Models}}
\label{sec:results_dl_benchmark}

Table~\ref{tbl:dl_benchmark_models} compares the configuration and study coverage of our LA3D approach and the related benchmark AN studies that employ DL models.

\begin{table}[!h] 
    \centering
    \caption{Deep Learning-based Benchmark AN Studies}
    \resizebox{1\columnwidth}{!}{
    \begin{tabular}{lccccc}
    \hline
    {Model Name} & {Complexity} & {AN Type} & {PD} & {ReID}   & {VAD / Action Recognition}  \\ \hline
    \textsc{SPAct}~\cite{dave2022spact} & DL  & Full-Image & \ding{51} & \ding{53} & \ding{51} \\ 
    \textsc{TeD-SPAD}~\cite{fioresi2023ted}  & DL & Full-Image & \ding{51} & \ding{53} & \ding{51}    \\ 
    \textsc{SG-GAN}~\cite{hukkelaas2023realistic} & DL  & Full-Body & \ding{53} & \ding{51} & \ding{53} \\
    \textsc{DeepPrivacy2}~\cite{hukkelaas2023deepprivacy2} & DL  & Full-Body & \ding{53} & \ding{51} & \ding{53} \\
    \textsc{LA3D} (ours) & Lightweight & Full-Body & \ding{51} & \ding{51} &  \ding{51} \\ \hline 
\end{tabular}
}
\label{tbl:dl_benchmark_models}
\vspace*{-0.5\baselineskip}
\end{table}

\subsection{{Further Evaluation on Person Re-Identification}}
\label{sec:results_reid_extra}

We have further evaluated ReID for recognition attacks that employ non-anonymized data on either the query or the search gallery. Table~\ref{tbl:reid_merket1501_clf_compare_all_aqrg} and Table~\ref{tbl:reid_merket1501_clf_compare_all_rqag} present the ReID protection capability of the ANs when the anonymized image of a person is utilized to search for identification from a gallery of raw images and vice versa, respectively. 
In both cases, the adaptive ANs consistently outperform the corresponding baselines. Blurring is more susceptible to ReID attack when both the query and gallery are anonymized, and the protection drops by approximately $20\%\text{--}26\%$ for $\mathcal{G}^\text{0}$, and $10\%\text{--}17\%$ for $\mathcal{G}^\text{a}$ in the mAP. The $\mathcal{G}^\text{a} _\text{max}$ is the least affected with decline of $10\%\text{--}11\%$.

\begin{table}[]
\centering
\caption{ReID (Anonymized Query vs. Raw Gallery) on the Market1501 Dataset~\cite{zheng2015scalable} using \textsc{OSNet}~\cite{zhou2021learning}.}
\resizebox{1\columnwidth}{!}{
\begin{tabular}{lcccccc}
\hline
AN Method                                          & mAP$\downarrow$ & $\Delta_\text{N}\downarrow$ & $\Delta_\text{0}\downarrow$ & CMC-R1$\downarrow$ & $\Delta_\text{N}\downarrow$ & $\Delta_\text{0}\downarrow$ \\ \hline
No-AN                                      & 0.826           & --                                  & --                              & 0.942              & --                                     & --                                     \\
MASKED                                       & 0.011           & -98.7\%                             & --                              & 0.010              & -98.9\%                                & --                                     \\
EDGED                                           & \textbf{0.009}  & \textbf{-98.9\%}                    & --                              & \textbf{0.008}     & \textbf{-99.2\%}                       & --                                     \\
$\mathcal{G}^\text{0}$                          & 0.165           & -80.0\%                             & --                              & 0.202              & -78.6\%                                & --                                     \\
$\mathcal{G}^\text{a}(\alpha_l=0.5)$            & 0.042           & -94.9\%                             & -74.5\%                         & 0.042              & -95.5\%                                & -79.2\%                                \\
$\mathcal{G}^\text{a} (\alpha_l=1.0)$           & 0.034           & -95.9\%                             & -79.4\%                         & 0.035              & -96.3\%                                & -82.7\%                                \\
$\mathcal{G}^\text{a} _\text{max}$ & 0.022           & -97.3\%                             & -86.7\%                         & 0.022              & -97.7\%                                & -89.1\%                                \\
$\mathcal{P}^\text{0}_2$                        & 0.800           & -3.1\%                              & --                              & 0.926              & -1.7\%                                 & --                                     \\
$\mathcal{P}^\text{0}_4$                        & 0.687           & -16.8\%                             & --                              & 0.832              & -11.7\%                                & --                                     \\
$\mathcal{P}^\text{0}_8$                        & 0.320           & -61.3\%                             & --                              & 0.413              & -56.2\%                                & --                                     \\
$\mathcal{P}^\text{a}_2(\alpha_l=0.5)$          & 0.079           & -90.4\%                             & -90.1\%                         & 0.092              & -90.2\%                                & -90.1\%                                \\
$\mathcal{P}^\text{a}_4(\alpha_l=0.5)$          & 0.071           & -91.4\%                             & -89.7\%                         & 0.079              & -91.6\%                                & -90.5\%                                \\
$\mathcal{P}^\text{a}_8(\alpha_l=0.5)$          & 0.073           & -91.2\%                             & -77.2\%                         & 0.079              & -91.6\%                                & -80.9\%                                \\
$\mathcal{P}^\text{a}_2(\alpha_l=1.0)$          & 0.074           & -91.0\%                             & -90.8\%                         & 0.085              & -91.0\%                                & -90.8\%                                \\
$\mathcal{P}^\text{a}_4(\alpha_l=1.0)$          & 0.060           & -92.7\%                             & -91.3\%                         & 0.059              & -93.7\%                                & -92.9\%                                \\
$\mathcal{P}^\text{a}_8(\alpha_l=1.0)$          & 0.027           & -96.7\%                             & \textbf{-91.6\%}                & 0.020              & -97.9\%                                & \textbf{-95.2\%}                       \\
$\mathcal{P}^\text{a} _\text{max}$ & 0.027           & -96.7\%                             & --                              & 0.020              & -97.9\%                                & --                                      \\ \hline
\end{tabular}
}
\label{tbl:reid_merket1501_clf_compare_all_aqrg}
\vspace*{-0.5\baselineskip}
\end{table}

\begin{table}[]
\centering
\caption{ReID (Raw Query vs. Anonymized Gallery) on the Market1501 Dataset~\cite{zheng2015scalable} using \textsc{OSNet}~\cite{zhou2021learning}.}
\resizebox{1\columnwidth}{!}{
\begin{tabular}{lcccccc}
\hline
AN Method                                          & mAP$\downarrow$ & $\Delta_\text{N}\downarrow$ & $\Delta_\text{0}\downarrow$ & CMC-R1$\downarrow$ & $\Delta_\text{N}\downarrow$ & $\Delta_\text{0}\downarrow$ \\ \hline

No-AN                                      & 0.826           & --                                  & --                              & 0.942              & --                                     & --                                     \\
MASKED                                       & 0.016           & -98.1\%                             & --                              & 0.096              & -89.8\%                                & --                                     \\
EDGED                                           & \textbf{0.013}  & \textbf{-98.4\%}                    & --                              & \textbf{0.093}     & \textbf{-90.1\%}                       & --                                     \\
$\mathcal{G}^\text{0}$                          & 0.217           & -73.7\%                             & --                              & 0.387              & -58.9\%                                & --                                     \\
$\mathcal{G}^\text{a}(\alpha_l=0.5)$            & 0.065           & -92.1\%                             & -70.0\%                         & 0.172              & -81.7\%                                & -55.6\%                                \\
$\mathcal{G}^\text{a} (\alpha_l=1.0)$           & 0.056           & -93.2\%                             & -74.2\%                         & 0.156              & -83.4\%                                & -59.7\%                                \\
$\mathcal{G}^\text{a} _\text{max}$ & 0.036           & -95.6\%                             & -83.4\%                         & 0.132              & -86.0\%                                & -65.9\%                                \\
$\mathcal{P}^\text{0}_2$                        & 0.806           & -2.4\%                              & --                              & 0.933              & -1.0\%                                 & --                                     \\
$\mathcal{P}^\text{0}_4$                        & 0.704           & -14.8\%                             & --                              & 0.895              & -5.0\%                                 & --                                     \\
$\mathcal{P}^\text{0}_8$                        & 0.350           & -57.6\%                             & --                              & 0.651              & -30.9\%                                & --                                     \\
$\mathcal{P}^\text{a}_2(\alpha_l=0.5)$          & 0.080           & -90.3\%                             & -90.1\%                         & 0.262              & -72.2\%                                & -71.9\%                                \\
$\mathcal{P}^\text{a}_4(\alpha_l=0.5)$          & 0.084           & -89.8\%                             & -88.1\%                         & 0.219              & -76.8\%                                & -75.5\%                                \\
$\mathcal{P}^\text{a}_8(\alpha_l=0.5)$          & 0.084           & -89.8\%                             & -76.0\%                         & 0.221              & -76.5\%                                & -66.1\%                                \\
$\mathcal{P}^\text{a}_2(\alpha_l=1.0)$          & 0.076           & -90.8\%                             & -90.6\%                         & 0.246              & -73.9\%                                & -73.6\%                                \\
$\mathcal{P}^\text{a}_4(\alpha_l=1.0)$          & 0.059           & -92.9\%                             & \textbf{-91.6\%}                & 0.193              & -79.5\%                                & -78.4\%                                \\
$\mathcal{P}^\text{a}_8(\alpha_l=1.0)$          & 0.032           & -96.1\%                             & -90.9\%                         & 0.126              & -86.6\%                                & \textbf{-80.6\%}                       \\
$\mathcal{P}^\text{a} _\text{max}$ & 0.032           & -96.1\%                             & --                              & 0.126              & -86.6\%                                & --                                      \\
*\textsc{SG-GAN}~\cite{hukkelaas2023realistic} & 0.144 & -82.6\% & -- & 0.311 &   -67.0\%    &  -- \\ 
*\textsc{DeepPrivacy2}~\cite{hukkelaas2023deepprivacy2} & 0.085 & -89.7\% & -- & 0.447 &   -52.5\%    &  -- \\ \hline
\end{tabular}
}
\noindent \par \footnotesize
The * are DL AN methods that employ body image generators. 
\par
\label{tbl:reid_merket1501_clf_compare_all_rqag}
\vspace*{-0.5\baselineskip}
\end{table}

\subsection{{Video Quality Assessment for Action Utility Preservation}}
\label{sec:results_iqa}

We present a video quality analysis to evaluate the general capability of preserving the action utility of the AN. The employed conventional ANs are perturbation-based, not aimed at generating a realistic image (like GANs), and the purpose is to retain the visual utility for the end target VAD task through the action recognition I3D video encoder. Hence, we employed quality metrics, such as {Fréchet Inception Distance} (FID) and {Kernel Inception Distance} (KID), to measure the closeness of retained information before and after AN on I3D-encoded features of the videos. 

\begin{figure}[]
\centering
\begin{subfigure}[]{0.9\columnwidth}
\centering
\includegraphics[width=\columnwidth]{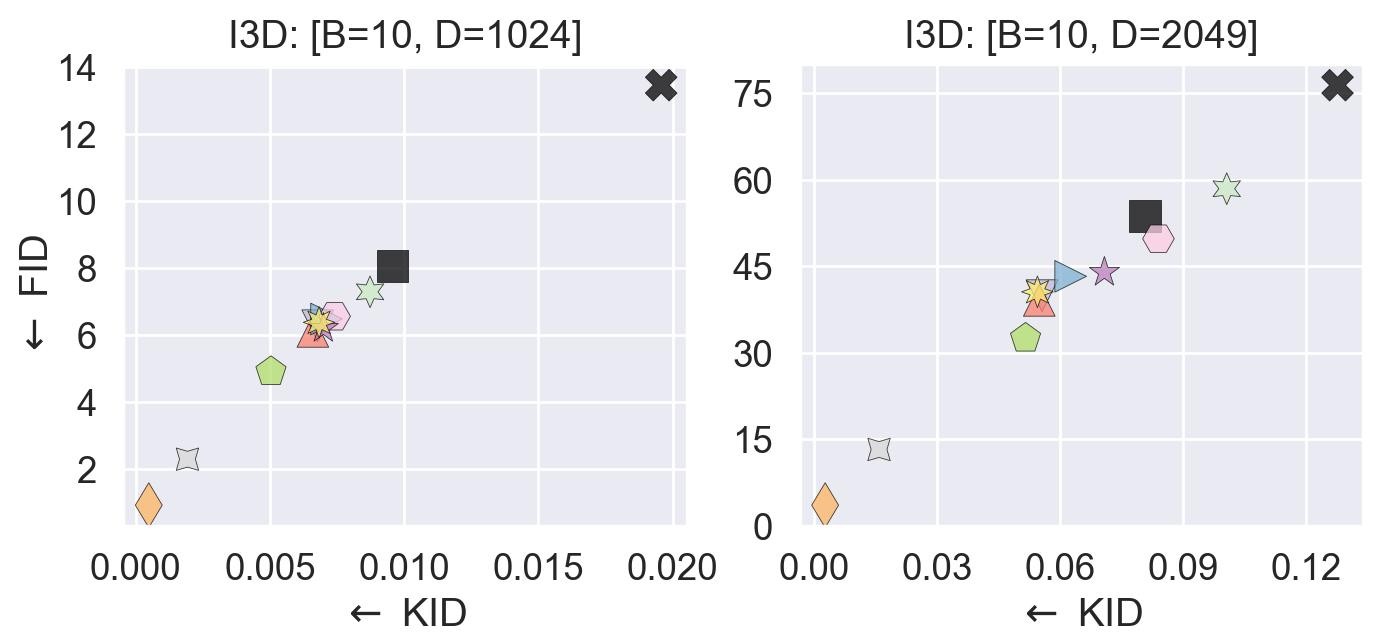} 
\caption{}
\end{subfigure}
\begin{subfigure}[]{0.9\columnwidth}
\centering
\includegraphics[width=\columnwidth]{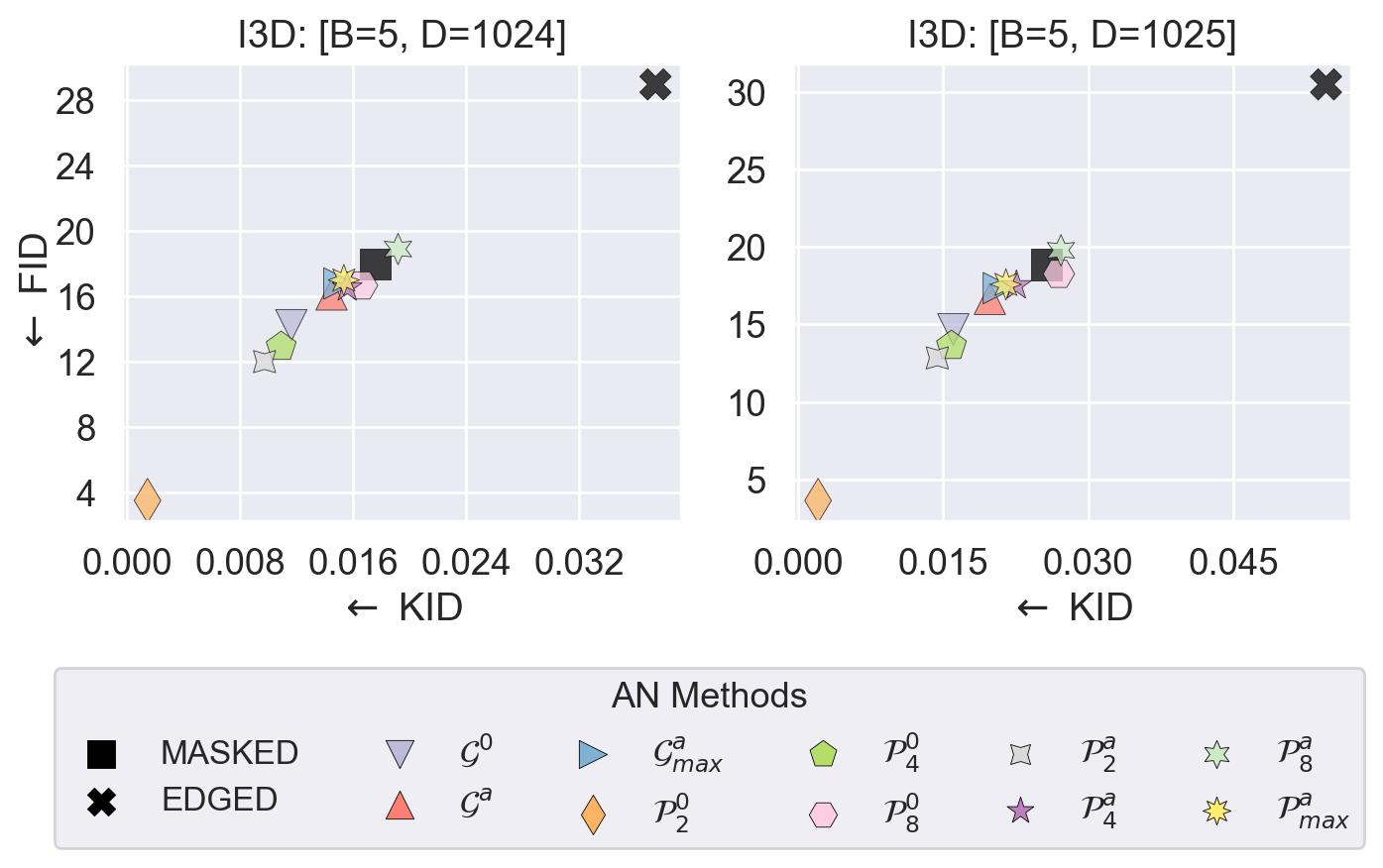} 
\caption{}
\end{subfigure}
\caption{Video quality evaluation of the anonymized videos: a) UCF-Crime dataset, and b) XD-violence. The FID and KID scores are computed on the extracted I3D features (with dimension of $B \times D$) of the videos, and the distance is measured from the I3D features of the No-AN videos. 
}
\label{fig:iqa_i3d_ucf_d1024_d2049_xd_d1024_d1025}
\vspace*{-\baselineskip}
\end{figure}

Fig.~\ref{fig:iqa_i3d_ucf_d1024_d2049_xd_d1024_d1025} presents the average quality scores over the 290 test videos of the UCF-Crime dataset and 800 videos of the XD-Violence.
The results overall correlate with the VAD performance, and the impact of AN exhibits consistency across the VAD and I3D models. However, the EDGED AN achieves the poorest quality scores. Despite these being in harmony with the VAD performance on the XD-Violence, the AN performed well for the MGFN on the UCF-Crime dataset (as shown in Fig.~\ref{fig:ad_ucf_auc_pd_vispr_cmap_clf_compare_all}). The adaptive ANs have provided results that are relatively lower than their baseline, which is expected considering the stronger AN. Nonetheless, the FID and KID scores are generally low, indicating a promising preservation of the action utilities. We hypothesize that the AN methods can lead to varying efficacy levels for non-VAD action recognition tasks, and an in-depth investigation for a given target task is essential.

\subsection{{Video Feature Processing for Video Anomaly Detection}}
\label{sec:vad_i3d_encoding}
 
The VAD models were trained on \textsc{I3D-RGB} extracted features of the raw videos of the UCF-Crime dataset with a frame size of [$320, 240$]. The \textsc{I3D} video encoder operates on an input video frame size of [$340, 256$] and generates features using a clip sequence length $V=16$, and crop augmentations $B=10$ with a crop size of [$224, 224$]. The \textsc{PEL4VAD} utilizes feature dimensions of $D=1024$ using \textsc{I3D} encoder from Ref.~\cite{carreira2017quo}, whereas \textsc{MGFN} adopts $D=2049$ using non-local \textsc{I3D} from Ref.~\cite{wang2018non}. 
 
Benchmark DL AN studies, such as \textsc{TeD-SPAD}~\cite{fioresi2023ted}, employ frame normalization for the \textsc{MGFN} that scales into the range of $[0, 1]$, where the value is divided by 255. However, we found that their reported VAD performance ($\text{AUC}=0.78$) is considerably lower than the reported scores by the original \textsc{MGFN} study in Ref.~\cite{chen2023mgfn} ($\text{AUC}=0.86$ with $V=32$). Following the implementation of Ref.~\cite{wang2018non}, we have achieved better scores ($\text{AUC}=0.83$) by standardizing the frames as:
\begin{equation}
    \mathbf{\bar{I}} = \frac{\mathbf{I} - \mu}{\sigma},
\end{equation}
where the $\mathbf{\bar{I}} \in \mathbb{R}$ is the standardized data of the input frame $\mathbf{I} \in \mathbb{Z}$, and the $\mu=114.75$ and $\sigma=57.375$ are image scaling mean and standard deviation factors, respectively, that are derived from the Kinetics400 dataset~\cite{kay2017kinetics,wang2018non}. 

We normalize the frames into $[-1,1]$ for \textsc{PEL4VAD} model using:
\begin{equation}
    \mathbf{\bar{I}} = \frac{2 \times \mathbf{I}}{255} - 1.
\end{equation}

For the XD-Violence dataset, the \textsc{I3D} encoder~\cite{carreira2017quo} operates on input video frames with a size of [$340, 256$] normalized into $[-1, 1]$. The encoder utilizes a clip sequence length $V=16$, and crop augmentations $B=5$ with a crop size of [$224, 224$]. Both the VAD models employ the same encoded features with a dimension of $D=1024$, but the \textsc{MGFN} incorporates one more feature through its feature amplification mechanism, which increases its final dimension to $D=1025$.

\subsection{{VAD Confidence Score Evaluation}}
\label{sec:results_vad_ad_score}

\begin{figure}[]
\centering
\begin{subfigure}[]{1\columnwidth}
\centering
\includegraphics[width=\columnwidth]{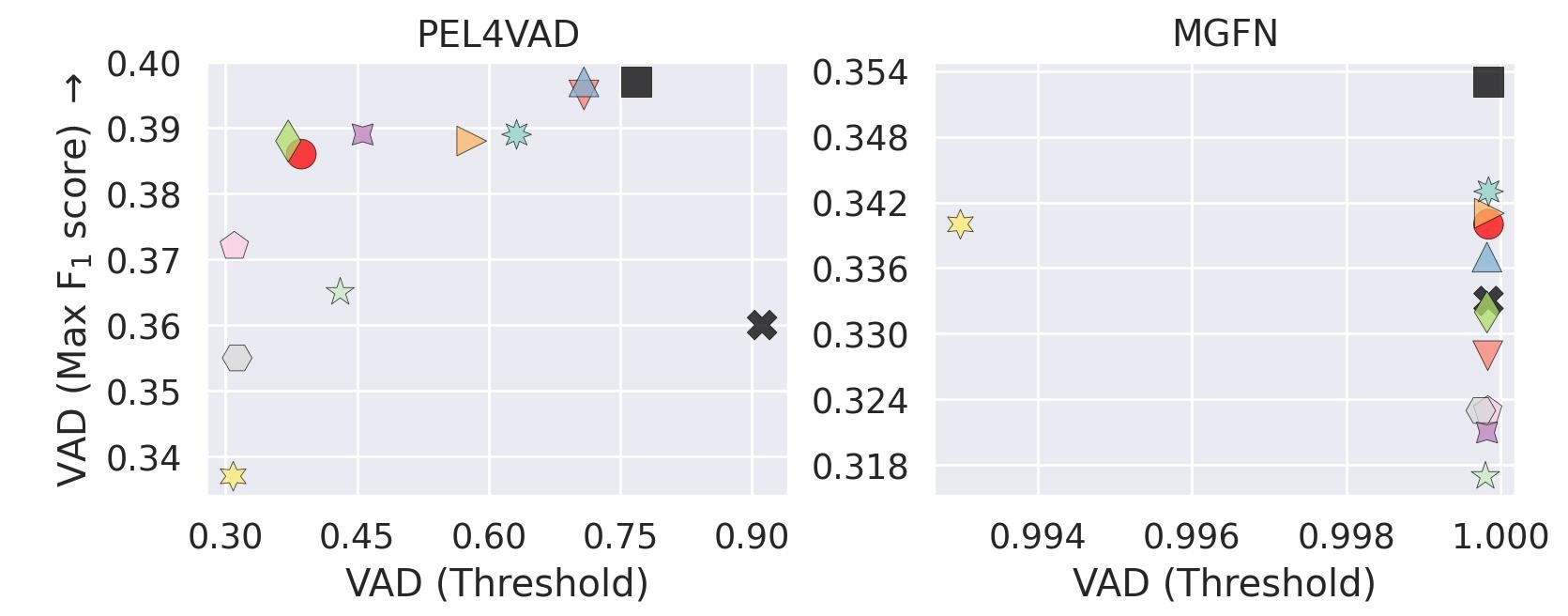} 
\caption{}
\end{subfigure}
\begin{subfigure}[]{1\columnwidth}
\centering
\includegraphics[width=\columnwidth]{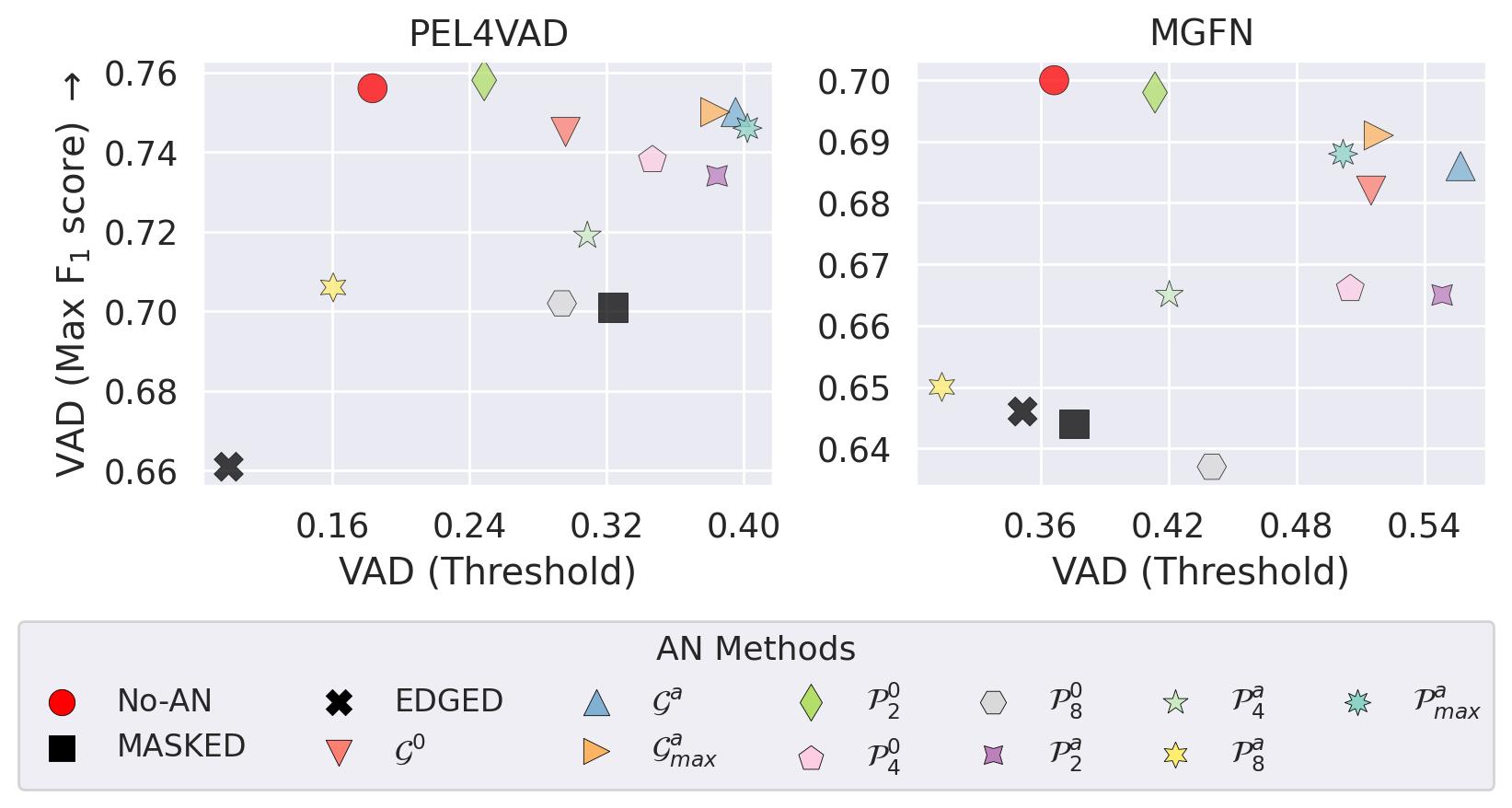} 
\caption{}
\end{subfigure}
\caption{VAD maximum $\text{F}_1$ scores and their thresholds: a) UCF-Crime dataset, and b) XD-violence. 
The plots highlight the need to adjust the detection threshold to achieve optimal performance when applying AN to a VAD.
}
\label{fig:PEL4VAD_MGFN_ad_f1_max_clf_compare_all}
\vspace*{-\baselineskip}
\end{figure}

This subsection presents an investigation of AN and VAD through temporal plots of anomaly confidence scores across video frames.
Fig.~\ref{fig:PEL4VAD_MGFN_ad_f1_max_clf_compare_all} depicts the best $\text{F}_1$ score along with its decision threshold for each AN method. Some of the AN methods, such as heavier $\mathcal{P}$, exhibit a lower confidence score and require a lower decision threshold when generating VAD flags, whereas others, such as MASKED and $\mathcal{G}$, need much higher thresholds. The plots demonstrate the necessity of adjusting thresholds when applying AN to a VAD for optimal rates of false positives and false negatives.

\begin{figure*}[!th]
\centering
\setlength{\kw}{1.4cm}
\begin{subfigure}[]{1\textwidth}
\centering
\resizebox{1\textwidth}{!}{
\begin{tabular}{p{0.5cm}K{\kw}K{\kw}K{\kw}K{\kw}K{\kw}K{\kw}K{\kw}K{\kw}K{\kw}}
& No-AN  & MASKED & EDGED & $\mathcal{G}^\text{0}$ & $\mathcal{G}^\text{a}$ ($\alpha_l=0.5$)  & $\mathcal{G}^\text{a}_\text{max}$ ($\text{AN}_\text{max}=1$) & $\mathcal{P}^\text{0}_4$ & $\mathcal{P}^\text{a}_4$ ($\alpha_l=0.5$) & $\mathcal{P}^\text{a}_\text{max}$ ($\text{AN}_\text{max}=1$)
\end{tabular}
}
\adjustimage{width=1\textwidth,center}{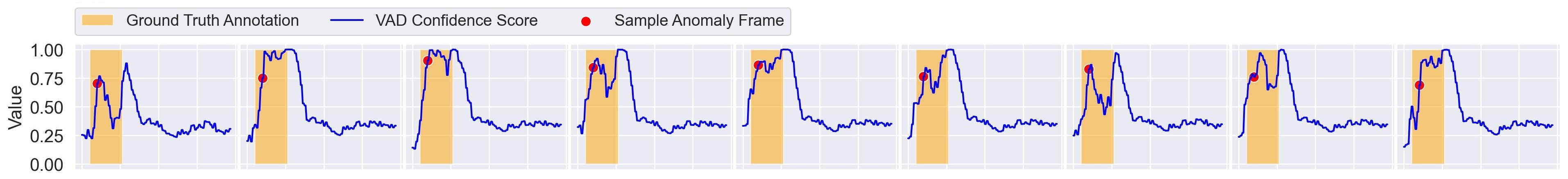}
\adjustimage{width=1\textwidth,center}{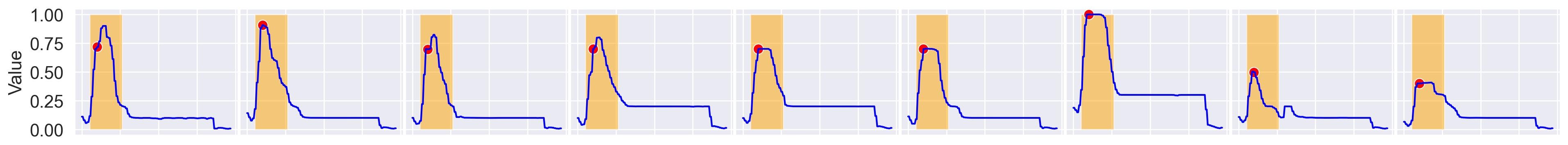}
\adjustimage{width=0.95\textwidth,right}{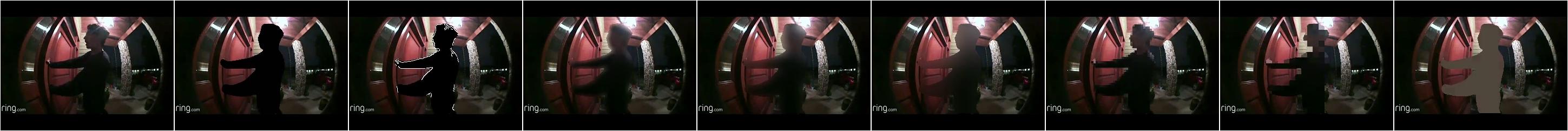}
\caption{UCF-Crime dataset: Burglary033\_x264.mp4}
\end{subfigure}
\begin{subfigure}[]{1\textwidth}
\centering
\adjustimage{width=1\textwidth,center}{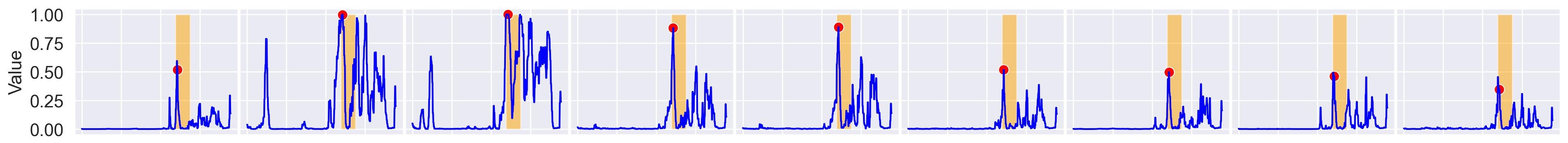}
\adjustimage{width=1\textwidth,center}{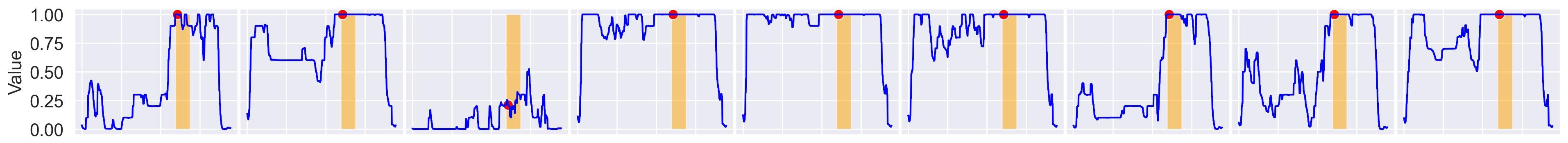}
\adjustimage{width=0.95\textwidth,right}{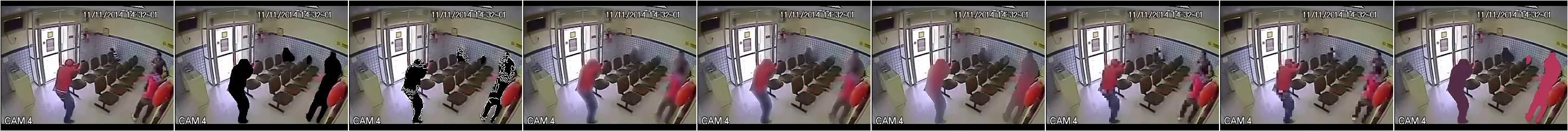}
\caption{UCF-Crime dataset: Shooting022\_x264.mp4}
\label{fig:ad_ucf_vis_sample_ad}
\end{subfigure}
\begin{subfigure}[]{1\textwidth}
\centering
\adjustimage{width=1\textwidth,center}{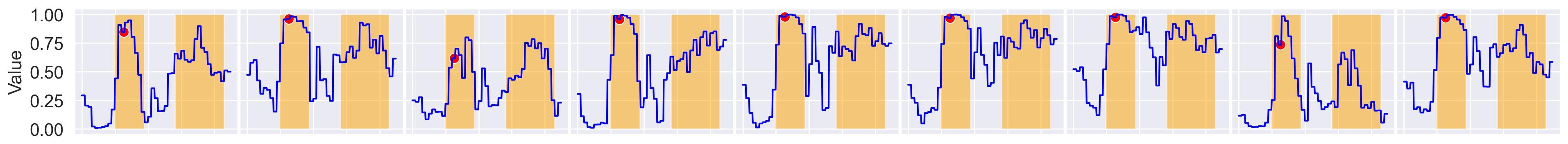}
\adjustimage{width=1\textwidth,center}{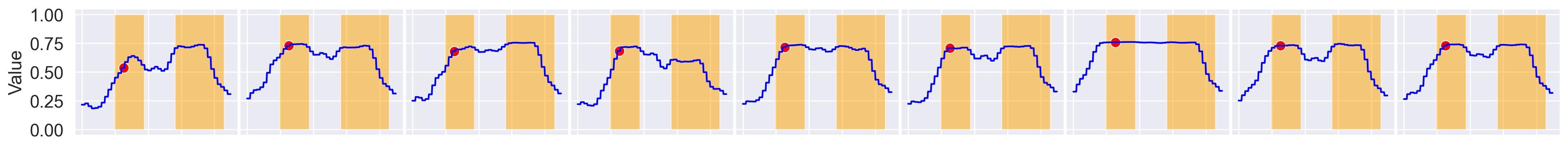}
\adjustimage{width=0.95\textwidth,right}{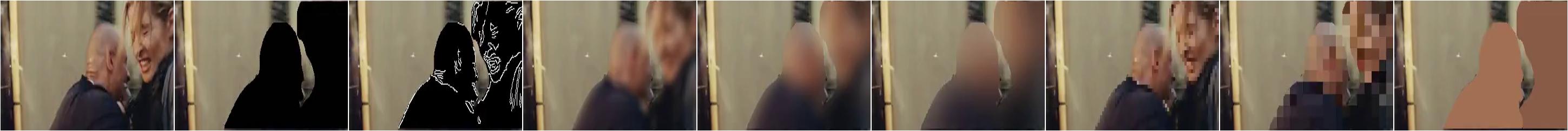}
\caption{XD-Violence dataset: Fast.Five.2011\_\_\#00-32-56\_00-33-26\_label\_B2-0-0.mp4}
\label{fig:ad_xd_vis_sample_ad}
\end{subfigure}
\begin{subfigure}[]{1\textwidth}
\centering
\adjustimage{width=1\textwidth,center}{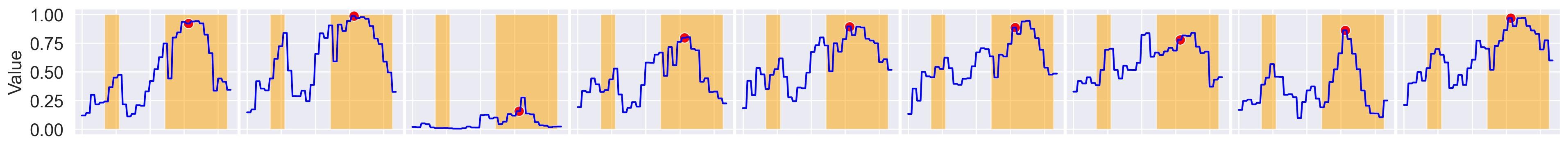}
\adjustimage{width=1\textwidth,center}{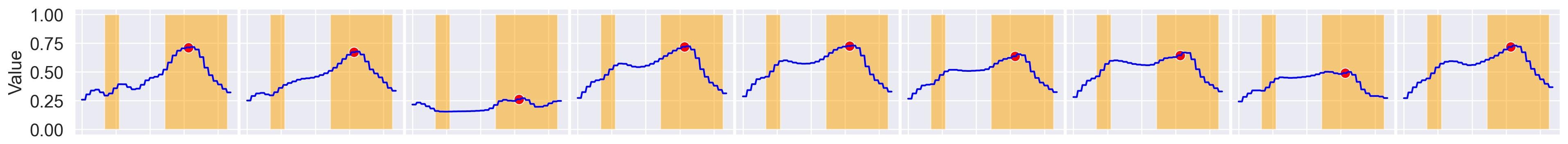}
\adjustimage{width=0.95\textwidth,right}{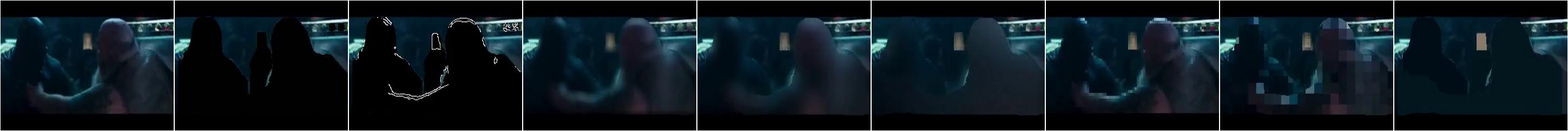}
\caption{XD-Violence dataset: Deadpool.2016\_\_\#0-18-58\_0-19-20\_label\_B1-0-0.mp4}
\end{subfigure}   
\caption{VAD anomaly score illustration on sample videos: (top) confidence score of the \textsc{PEL4VAD}~\cite{pu2024learning}, (middle) confidence score of the \textsc{MGFN}~\cite{chen2023mgfn}, and (bottom) video frame sampled at the red dot markers of the score plots). 
The \textsc{PEL4VAD} has provided better anomaly localization across the AN methods, whereas the \textsc{MGFN} has struggled to localize in (b).
Both VAD models have achieved good anomaly localization in (c) and (d), with \textsc{PEL4VAD} offering superior separation of the anomaly regions. The EDGED AN limits the performance in (d) for both models. 
}
\label{fig:ad_ucf_xd_vis_sample_ad}
\vspace*{-1\baselineskip}
\end{figure*}

Figs.~\ref{fig:ad_ucf_vis_sample_ad} and.~\ref{fig:ad_xd_vis_sample_ad} portray a visual illustration of the VAD confidence scores of the AN methods on sample videos from the UCF-Crime and XD-Violence datasets, respectively. The \textsc{PEL4VAD} has demonstrated promising anomaly localization across the AN methods, while the \textsc{MGFN} struggles with some of the videos. The figures also show that different AN methods have varying impacts on the strength of the anomaly scores. Methods, including the MASK, EDGED, and $\mathcal{G}$, have an increasing effect, whereas the $\mathcal{P}$ can diminish the scores. 

\subsection{{Limitations and Future Research Directions}}
\label{sec:results_limitation}

Despite the encouraging performance of the proposed approaches, we outline potential limitations and future study considerations below:
\begin{itemize}
    \item \textit{Optimization of hyperparameters}: The adaptive $\Theta^\text{a}$ initializes with the base $\Theta^\text{0}$. This may lead to varying performance depending on the choice of the base parameters of $\Theta^\text{0}$, e.g., the varying capacity of $\mathcal{P}^\text{a}_{d \in \{2,4,8\}}$ on the AN and VAD. We have found $\mathcal{G}^\text{0}_{k \in \{10, \dots, 15\}}$ and $\mathcal{P}^\text{0}_4$ provide good base parameters at image size of $\mathbf{z_\text{ref}}=[320, 240]$ with the scaling factor of $\alpha_r = \mathbf{z}/\mathbf{z_\text{ref}}$ for larger resolutions. Although the $\text{AN}_\text{max}=1$, that enforces the maximum adaptive AN, avoids the dependency on the hyperparameters, it may deteriorate the AN for $\mathcal{P}^\text{a}_\text{max}$ and increases the computational cost for $\mathcal{G}^\text{a}_\text{max}$. Thus, we recommend further study in auto-optimization of base parameters to enhance performance. 
     
    \item \textit{Target depth vs. surface area}: We have considered the segmentation mask area as a rough approximation of depth, i.e., entities at shallow depth will have higher areas and vice versa. Although this performs in most cases, occluded entities violate this assumption. Depth estimation models offer more accurate assessments of object depth in images, thereby enhancing strategies for addressing this challenge. Another potential solution is to employ dense pose estimation, a method that helps segment human body parts and has been utilized for generating synthetic images~\cite{guler2018densepose}. This technique can also assist in detecting occlusions by identifying missing body parts. However, the implementation of these DL approaches in computationally limited and real-time settings is often constrained due to their slow processing speeds and limited accuracy~\cite{hukkelaas2023deepprivacy2} (see Fig.~\ref{fig:cc_vispr_sample_320_240}). We have tested lightweight depth estimation models, such as \textsc{Lite-Mono} models~\cite{zhang2023lite}, and found their accuracy inadequate for depth estimation of human subjects; they are excessively sensitive to human skin and clothing. Our study has attempted to address the challenge using damped scaling for decent occlusions (see Eq.~\eqref{eq:adaptive_r}). We have also introduced $\text{AN}_\text{max}=1$, which ensures the maximum AN with relatively significant computational leverage. But, it would result in non-uniform AN for targets at the same depth with varying areas due to occlusion. 
    For future research work on improved adaptive AN methods, we recommend integrating enhanced lightweight depth estimation with the surface area of target subjects. For instance, the adaptive scaling factor $r$ in Eq.~\eqref{eq:adaptive_r} can further be weighted with a depth score.
     
    \item \textit{Miss-detections}: We have employed an image-level segmentation, via the YOLO medium model, as a high-accuracy human detector. But, some targets can still be missed due to lower image quality, such as occlusion, motion, lighting, weather, and image resolution. Object segmentation study is a separate domain, and further analysis on it is beyond the scope of the AN research. Nevertheless, we recommend using larger YOLO models with higher accuracy, employing data augmentation, or utilizing video-level detectors with tracking capability to fill the gap and improve detection with additional overhead~\cite{duke2021sstvos, cheng2022xmem, shi2024practical}.
\end{itemize}

\end{document}